\documentclass[11pt]{article}
\usepackage[final]{acl}
\usepackage{times}
\usepackage{latexsym}
\usepackage[T1]{fontenc}
\usepackage[utf8]{inputenc}
\usepackage{microtype}
\usepackage{inconsolata}
\usepackage{graphicx}
\usepackage{amsmath}
\usepackage{amssymb}
\usepackage{booktabs}
\usepackage{multirow}
\usepackage{array}
\usepackage{xcolor}

\newcommand{\pp}{\,\text{pp}}
\newcommand{\van}{\textsc{base}}
\newcommand{\sft}{\textsc{distilled}}
\newcommand{\teach}{\textsc{teacher}}
\newcommand{\weakt}{\textsc{weak-t}}
\newcolumntype{L}[1]{>{\raggedright\arraybackslash}p{#1}}

\title{Better Accuracies, Worse Reasoning:\\A Step-Level Audit of Medical Chain-of-Thought Distillation}

\author{
  \textbf{Zhaoyang Jiang\textsuperscript{1}},
  \textbf{Xuanqi Peng\textsuperscript{1}},
  \textbf{Fei Teng\textsuperscript{2}},
  \textbf{Zhizhong Fu\textsuperscript{3}},
\\
  \textbf{Yunsoo Kim\textsuperscript{4}},
  \textbf{Jiacong Mi\textsuperscript{1}},
  \textbf{Zicheng Li\textsuperscript{1}},
  \textbf{Honghan Wu\textsuperscript{1}}
\\
\\
  \textsuperscript{1}School of Health \& Wellbeing, University of Glasgow, Glasgow, UK,
\\
  \textsuperscript{2}Department of Respiratory and Critical Care Medicine, Shanghai Sixth People's Hospital,
\\
  Shanghai Jiao Tong University School of Medicine, Shanghai, China,
\\
  \textsuperscript{3}School of Life Science and Technology,
\\
  University of Electronic Science and Technology of China, Chengdu, China,
\\
  \textsuperscript{4}Institute of Health Informatics, University College London, London, UK
\\
  \small{
    \textbf{Correspondence:} \href{mailto:Honghan.Wu@glasgow.ac.uk}{Honghan.Wu@glasgow.ac.uk}
  }
}

\begin{document}
\maketitle

\begin{abstract}
Chain-of-thought (CoT) distillation trains a smaller model to imitate a teacher's reasoning trace, but it is typically evaluated by final-answer metrics. We ask whether gains in answer quality are accompanied by improvements in the trace. In medical QA, where short answer options can leave a richer clinical justification under-specified, a Qwen3-8B student distilled from a DeepSeek-V3-family teacher improves on MedQA-USMLE answer metrics (SC@64 $74.7\%\!\to\!84.1\%$; expected calibration error (ECE) $0.096\!\to\!0.034$). Yet under a style-blind Kimi-K2.6 judge, its per-question factual step-error rate rises by $+16.1$ percentage points on average (paired 95\% CI $[14.5, 17.7]$), and two clinical experts independently labelling 450 blinded steps each find a gap of about $+13$ points. In this primary medical setting, answer quality and trace factuality move in opposite directions. This before--after pattern persists across evaluators, teacher strengths, student scales and families, medical benchmarks, and style, segmentation, and answer-correctness controls. Boundary checks narrow the claim: the risk appears when a compact answer under-constrains the rationale and a capable student can imitate expert-like form without reliably grounding each local claim. Standard answer metrics and aggregate hedging rates do not reveal the shift. When such traces are released or reused, answer-level metrics alone are insufficient.

\end{abstract}

\section{Introduction}

The modern recipe for making a compact reasoning model is simple: ask a stronger teacher to write chain-of-thought solutions, then train the smaller student to imitate them \citep{magister2023teaching,ho2023large,hsieh2023distilling,deepseekr1}. If the student's task-level output improves, and if its confidence looks better calibrated on multiple-choice benchmarks, the distillation is usually counted as a win.

\begin{figure*}[t]
\centering
\includegraphics[width=1\textwidth]{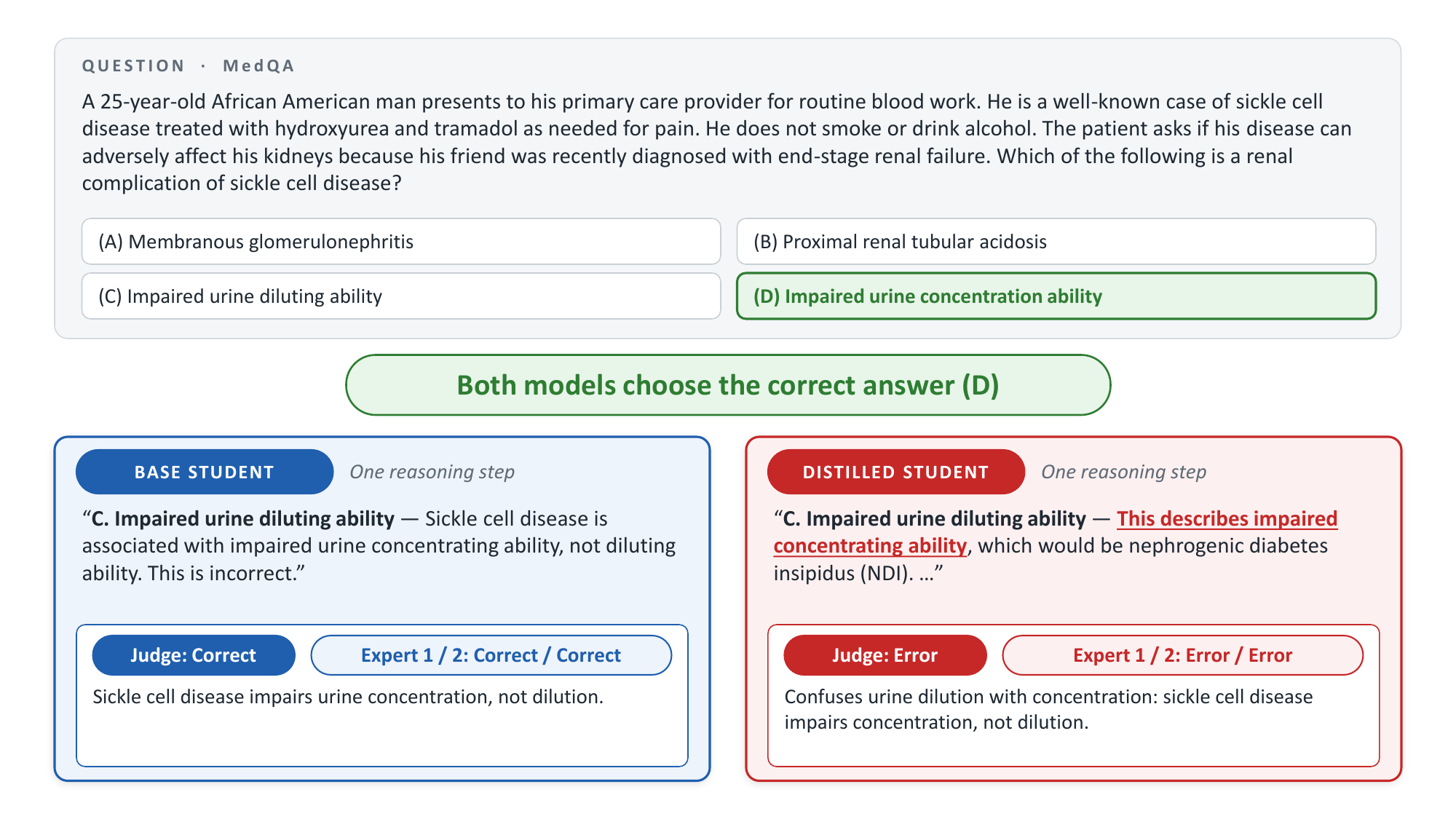}
\caption{An example from the MedQA test set. Both students choose the gold answer (D) and write a step evaluating option C. The base step is correct; the distilled step equates impaired diluting ability with impaired concentrating ability, an error flagged by the Kimi-K2.6 judge and by both clinical experts.}
\label{fig:overview}
\end{figure*}

That scoring assumes that the final output is the only product. Medicine makes the assumption concrete. A final answer in medical QA is a low-bandwidth target; the rationale is not. It states symptoms, mechanisms, exclusions, and differential diagnoses, any of which may matter to a reader even when the final answer is correct. A concrete example is cognitive diagnosis: distinctions among normal cognition, mild cognitive impairment, and Alzheimer-type dementia depend on judgment, functional history, test evidence, and exclusionary causes \citep{petersen1999mild,albert2011diagnosis,mckhann2011diagnosis,jiang2026lov3d}. Even the endpoint label can require substantial adjudication. In one consensus-diagnosis study, $298/635$ cases required conference calls, independent raters agreed with the final label $78.3\%$ of the time on average ($\kappa=0.50$), and agreement was lowest for MCI ($\kappa=0.34$) \citep{bassil2023feasibility}. When the label is not self-sufficient, the rationale becomes part of what a reader must inspect: it shows which evidence was used, which alternatives were excluded, and why one compact answer was preferred.

That is precisely the expensive object to verify. Medical-domain experts who can check step-level claims are scarce, which makes teacher-written CoTs attractive supervision. But the trace is not decoration. It is a justification that a medical reader may inspect, the surface CoT-monitoring methods hope to use for unsafe behavior \citep{korbak2025chainofthought,baker2025monitoring}, and the raw material for datasets that verify steps rather than merely answers \citep{wu2025medreason,sun2025reasonmed}. A trace can therefore improve the final answer while becoming less trustworthy as a sequence of factual claims.

This paper asks whether that failure mode appears in medical CoT distillation. We distill Qwen3-8B \citep{qwen3} with LoRA \citep{hu2022lora} from a DeepSeek-V3-family API teacher \citep{deepseekv3} on MedQA-USMLE \citep{jin2021disease}, then audit the generated traces at the step level. Each step is judged in isolation under a style-blind protocol: the judge sees the question, answer key, and one step, and is instructed to ignore fluency, tone, and verbal confidence and label only factual correctness. The audit evaluates the written rationale itself, the part of the distilled model that readers actually see.

Figure~\ref{fig:overview} illustrates the split on one MedQA question: both students reach the gold answer, but only the distilled step misstates the underlying physiology. By the usual answer-level metric suite, distillation looks successful: greedy accuracy rises by $3.4\pp$, self-consistency accuracy from $74.7\%$ to $84.1\%$, expected calibration error (ECE) falls from $0.096$ to $0.034$, and the Brier score falls from $0.158$ to $0.106$. ECE and the Brier score are calibration metrics: both are lower when a model's confidence matches its actual correctness, so the distilled model also appears better calibrated. Yet under our primary judge, the per-question rate of steps flagged as factually erroneous rises by $+16.1\pp$ on average (paired $95\%$ CI $[+14.5,+17.7]\pp$; pooled $30.6\%\!\to\!50.3\%$), and two clinical experts reviewing 450 blinded steps find gaps of about $+13\pp$. The before--after direction persists across judge configurations and under changes to teacher strength, student scale, student family, and medical benchmark.

Our contribution is a process-level diagnosis: in a standard medical CoT-distillation recipe, answer accuracy and calibration improve while the factual reliability of the written reasoning trace degrades sharply, and that degradation is hidden from the metrics normally reported for distilled reasoning models. MedMCQA and MedBullets5 show that the split is not confined to one medical test set. Nonmedical checks instead locate the boundary. The risk appears when a compact answer weakly constrains a richer rationale, the student can imitate the teacher's explanatory form, and local claims require domain knowledge that the answer label does not verify. Student-side checks sharpen the same boundary. Larger Qwen3 students (14B and 32B) preserve the split with smaller but significant step-error increases, and Llama-3.1-8B, another capable 8B base, repeats the MedQA degradation. Mistral-7B starts much lower in answer accuracy and much higher in step error; for that weaker base, teacher-trace SFT supplies missing task competence and improves judged steps instead of producing expert-like but poorly grounded rationales. 

The code used in this study is publicly available at \url{https://github.com/ZHAOYANG0705/medical-cot-distillation-audit}.

\section{Related Work}

CoT distillation improves small-model performance across many reasoning settings \citep{magister2023teaching,ho2023large,hsieh2023distilling,ranaldi2024aligning,deepseekr1}. The closest critique is output-level: \citet{wadhwa2024investigating} show that gains can survive shuffled rationales, rationale-after-label orderings, and reductions to a few key tokens, suggesting that the rationale signal need not be coherent reasoning. Our question is complementary. Holding the student fixed before and after one deployed distillation recipe, we ask whether the intermediate steps themselves become less factually reliable.

That question matters because a large body of process-supervision work treats steps as objects worth supervising or rewarding \citep{lightman2024lets,uesato2022solving}, and recent medical-reasoning corpora verify reasoning at the step level for precisely this reason \citep{wu2025medreason,sun2025reasonmed}. It also connects to the faithfulness and monitoring literatures: CoT explanations can be unfaithful to the computation that produced the answer \citep{turpin2023language,lanham2023measuring,paul2024making,stechly2024chain}, and CoT monitoring has been proposed as a safety mechanism but also flagged as fragile \citep{korbak2025chainofthought,baker2025monitoring}. Adjacent work shows that CoT prompting can hurt clinical-task accuracy \citep{wu2025whycot}, and that reasoning-tuned model families can hallucinate more on factuality benchmarks \citep{yao2025reasoning}. We isolate a different object: the same student model before and after a specific CoT-distillation intervention, evaluated at the factuality of individual trace steps.

This framing does not require the trace to be a faithful causal explanation of the model's computation. If a user, judge, or future training run consumes the emitted trace as text, the relevant question is whether distillation makes that text more or less factually dependable. Even an unfaithful trace can be harmful if it is read, audited, or reused as a medical rationale. Conversely, a trace can help the final answer while becoming a worse object for those downstream uses.

Our diagnosis also draws on work on verbalised uncertainty and internal representations. Models can be trained to express uncertainty in words \citep{lin2022teaching,mielke2022reducing}, but those words can become miscalibrated \citep{tian2023just,zhou2024relying}, and internal truth-related signals need not surface faithfully in text \citep{kadavath2022language,azaria2023internal,burns2022discovering}. We show a similar trace-level split: hedging remains informative inside a chain but fails to track the reliability shift caused by distillation. Our evaluation follows work on LLM-judge variability and style confounds \citep{zheng2023judging,llmjudgesurvey,ye2025justice,wu2025style}; our diagnostic activation analysis follows activation-patching and representation-editing practice \citep{zhang2024towards,arditi2024refusal,elhage2021mathematical}.

\section{A Controlled Audit Setting}
\label{sec:setup}

We use a single teacher--student pipeline for the primary audit, so that the before--after comparison has minimal experimental drift. The teacher is a DeepSeek-V3.2 API model \citep{deepseekv3}; the student is Qwen3-8B \citep{qwen3}. We generate teacher CoT solutions for MedQA-USMLE training questions \citep{jin2021disease} and LoRA-fine-tune \citep{hu2022lora} the student on them. We call the original student $\van$, the distilled student $\sft$, and the teacher's own traces $\teach$. As a teacher-strength control, we train a second Qwen3-8B student on CoTs from a local Qwen3-14B model with thinking disabled; this condition is denoted \weakt{}, with ``weak'' defined relative to the DeepSeek-V3.2 teacher. Task-level evaluation uses the 1{,}273-question MedQA-USMLE test set with greedy accuracy and SC@64. We repeat the same student-side check with Qwen3-14B, Qwen3-32B, and Llama-3.1-8B, and use Mistral-7B as a weak-base boundary. Medical transfer audits use MedMCQA \citep{pal2022medmcqa} and MedBullets5 \citep{chen-etal-2025-benchmarking}; GSM8K \citep{cobbe2021training}, MATH \citep{hendrycks2021math}, and ARC-Challenge \citep{clark2018think} are boundary diagnostics rather than cross-domain replications.

The teacher runs in non-thinking mode at temperature $0.7$, and each of the $10{,}178$ MedQA training questions contributes two traces. Traces come from two rounds. The teacher first answers without seeing the key, and we keep traces whose final answer is correct. Questions on which the teacher cannot produce enough correct traces this way receive the remaining traces with the correct option shown, and the teacher is asked to reason toward it. Of the resulting $20{,}356$ traces, $870$ ($4.3\%$) are answer-hinted ($11.0\%$ for \weakt{}); Section~\ref{sec:cost} reports a retrain without them. The primary student, its weaker-teacher and unhinted controls, and the larger Qwen3 students share one recipe: LoRA (rank $16$, $\alpha{=}32$) on all attention and MLP projections, three epochs at learning rate $2{\times}10^{-5}$, with the loss computed only on the trace. App.~\ref{app:training} lists the full settings.

Medical QA is useful here not because multiple-choice exams exhaust clinical reasoning, but because they separate the answer from the written process. The answer key anchors the final-output metric; the chain contains claims that are not reducible to the option letter. This lets us measure the object the benchmark normally ignores.

The audit target is narrower than ``is this a good explanation?'' and more concrete than faithfulness. A step can be causally irrelevant to the answer and still be factually wrong; it can also point toward the right option while misstating the medical relation that makes the option plausible. We judge the trace at that textual level: does the emitted step contain a factual error, given the case and the answer key? We are asking whether the step is safe to read, inspect, or reuse, not whether the model internally used it.

For the process-level audit, we take one sampled CoT per question, segment it with a fixed rule, and judge each step in isolation as \texttt{correct}, \texttt{error}, or \texttt{uncertain}. The judge sees the question, answer key, and one step, but not the full chain, so it cannot reward a globally plausible narrative. We use the first of the 64 self-consistency samples for the main audit and re-audit additional samples as a stability check. The prompt, segmentation rule, and audited-chain source are in App.~\ref{app:audit}.

The primary prompt is style-blind: it explicitly asks the judge to ignore tone, fluency, and confidence markers and to evaluate only factual content. Our primary judge is Kimi-K2.6, run over the full MedQA test set for each model ($7$--$11$k labelled steps each, $4$--$9\%$ \texttt{uncertain}). For robustness, we also run GLM-4-32B with both style-blind and naive prompts, and Hunyuan-A13B with the style-blind prompt, on a random 500-question subset. The step-error rate is $n_{\mathrm{err}}/(n_{\mathrm{err}}+n_{\mathrm{corr}})$ over steps labelled \texttt{error} or \texttt{correct}, excluding \texttt{uncertain}. This is a judge-conditioned committed-error estimate, not a style score or a ground-truth prevalence claim. The main evidential burden is the paired before--after increase and its stability across judges, prompts, and controls.

\section{Results}
\label{sec:cost}

Table~\ref{tab:medqa_summary} gives the overall result summary. The first row is the primary controlled comparison: on the 1{,}273-question MedQA test set, distillation improves accuracy and calibration while the audited trace worsens. Greedy accuracy rises from $72.9\%$ to $76.3\%$, SC@64 rises from $74.7\%$ to $84.1\%$, ECE falls from $0.096$ to $0.034$, and Brier score falls from $0.158$ to $0.106$. Taken alone, this metric suite would suggest an unambiguous improvement.

\begin{table*}[t]
\centering
\scriptsize
\setlength{\tabcolsep}{2pt}
\renewcommand{\arraystretch}{1.18}
\begin{tabular*}{\textwidth}{@{\extracolsep{\fill}}L{0.16\textwidth}L{0.18\textwidth}L{0.17\textwidth}L{0.22\textwidth}L{0.17\textwidth}@{}}
\toprule
Evidence source & Medical setting & Answer change & Reasoning-error change & Interpretation \\
\midrule
Primary audit & MedQA, Qwen3-8B & SC@64 $74.7{\to}84.1$; ECE/Brier down & $30.6{\to}50.3$; teacher $17.3$ & accuracy improves, reasoning error rises \\
Student scale/family & MedQA-500:\newline Qwen3-14B/32B; Llama-3.1-8B & Qwen3-14B SC@64 $78.9{\to}86.8$\newline Qwen3-32B 1-chain $81.8{\to}84.2$\newline Llama $66.8{\to}73.6$ & Qwen3-14B $31.5{\to}37.9$\newline Qwen3-32B $22.8{\to}30.6$\newline Llama $31.2{\to}45.5$ & capable students repeat \\
Weaker teacher & MedQA, Qwen3-8B & $74.7{\to}78.8$ ($+4.1\pp$) & $30.6{\to}41.1$ & smaller teacher signal, smaller cost \\
Medical OOD & MedMCQA, MedQA-SFT Qwen3-8B & $59.6{\to}59.4$ ($-0.2\pp$) & $32.2{\to}49.5$ & reasoning cost without task gain \\
Clinical vignette & MedBullets5, MedQA-SFT Qwen3-8B & $53.4{\to}57.4$ ($+4.0\pp$); teacher $76.5$ & $65.5{\to}78.4$; teacher $44.3$ & clean teacher, noisier student reasoning \\
Human audit & 450 blinded medical steps; two clinical experts & n/a & E1: $22.8{\to}35.8$; T $4.4$\newline E2: $23.7{\to}36.6$; T $8.3$ & two-expert check \\
\bottomrule
\end{tabular*}
\caption{Medical evidence for the accuracy--reasoning split. Answer metrics follow each setting's evaluation protocol. Automated step/segment-error rates are pooled Kimi-K2.6 style-blind estimates; human-audit rates are post-stratified. All error rates exclude \texttt{uncertain}. E1/E2 denote experts 1/2; T denotes the teacher.}
\label{tab:medqa_summary}
\end{table*}

The step-level audit reverses that conclusion. The paired per-question mean increase in step error is $+16.1\pp$ with bootstrap $95\%$ CI $[+14.5,+17.7]\pp$; a paired Wilcoxon signed-rank test over the $1{,}086$ questions with at least one non-\texttt{uncertain} step in both models gives $p<10^{-65}$ (cluster-robust $z=18.3$). The teacher's own full-test traces are cleaner at $17.3\%$. The weaker-teacher control lands in between at $41.1\%$, about half the inflation. The ordering is therefore not a generic effect of supervised fine-tuning alone: under the same judge and segmentation protocol, a weaker teacher produces a smaller cost.

Part of the primary gap comes from a small slice of the training data. The $4.3\%$ of teacher traces generated with the answer shown (Section~\ref{sec:setup}) are, by construction, explanations written toward a known answer. We retrain the student without them, using the same recipe, and audit its first chain on the first 500 MedQA questions. On these questions, the paired gap falls from $+16.0\pp$ for the primary student to $+11.8\pp$ ($95\%$ CI $[+9.4,+14.1]\pp$, $n=499$; pooled $31.0\%\!\to\!46.1\%$). The hinted traces therefore account for about a quarter of the primary gap ($+4.2\pp$, $95\%$ CI $[+2.0,+6.4]\pp$), but the split does not depend on them: the unhinted student's first-chain accuracy is $76.0\%$, against $73.4\%$ for the base. Nor do they explain the teacher-strength ordering, since the weaker-teacher data contain a larger hinted share ($11.0\%$) yet produce the smaller cost.

Larger Qwen3 students show the same split. A Qwen3-14B student trained with the same recipe improves MedQA SC@64 from $78.9\%$ to $86.8\%$. On a 500-question step audit, its pooled committed-error rate rises from $31.5\%$ to $37.9\%$; averaged by question, the paired mean increase is $+3.9\pp$ with $95\%$ CI $[+1.6,+6.4]\pp$ (Wilcoxon $p=8.9{\times}10^{-4}$). A Qwen3-32B check gives the same sign: first-chain accuracy rises from $81.8\%$ to $84.2\%$, while pooled step error rises from $22.8\%$ to $30.6\%$ (paired mean $+5.2\pp$, $95\%$ CI $[+3.0,+7.4]\pp$, Wilcoxon $p=1.9{\times}10^{-6}$). The effects are smaller than for Qwen3-8B, but they are not an 8B-only artifact.

We do not treat the Kimi percentages as calibrated prevalence estimates. They are committed-error estimates under one strong judge, and the panel below shows that different judges set different absolute baselines. The claim that survives the audit design is paired and comparative: with the same questions, the same step splitter, the same prompt, and the same judge, the distilled trace is flagged more often. A step may contain multiple claims and is flagged as erroneous if any one is judged factually incorrect, for example when it rules out a disease solely because a common but nonessential symptom is absent. The reported percentages therefore measure steps flagged as containing an error, rather than the proportion of individual medical claims that are false. The more stable fact is the before--after movement of the same student under the same measurement procedure.

Conditioning on the audited chain's final answer does not erase the split. Among the $784$ questions where both $\van$ and $\sft$ choose the correct option in that chain, committed step-error rates are $22.1\%$ vs.\ $37.5\%$; the paired question-level gap is $+12.1\pp$ with $95\%$ CI $[+10.2,+14.1]\pp$ (Wilcoxon $p=1.9{\times}10^{-30}$). Thus even a correct-answer filter would still admit substantially noisier distilled rationales. App.~\ref{app:answer_conditioned} gives the conditioning details.

\begin{figure*}[t]
\centering
\includegraphics[width=\textwidth]{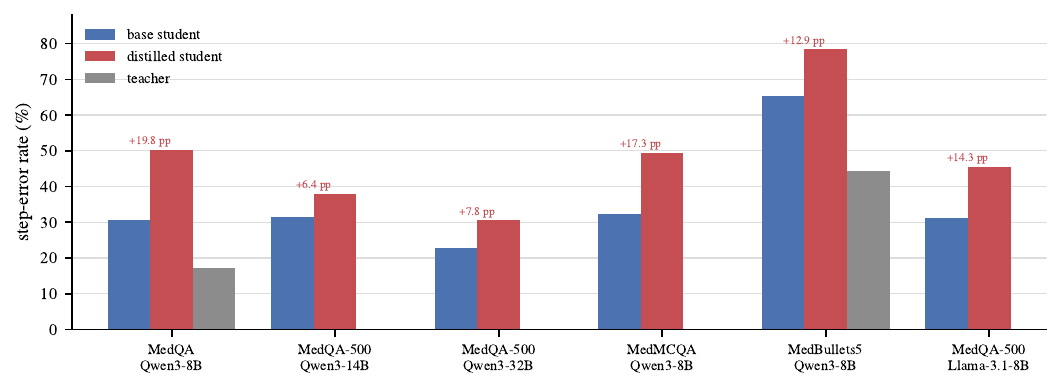}
\caption{Medical positive evidence. Distilled reasoning traces are flagged more often than base traces on MedQA, the Qwen3-14B and Qwen3-32B same-family scale checks, two clinical-QA transfer checks, and a second capable student family. Bar labels give pooled distilled-minus-base gaps; the paired per-question MedQA gap is $+16.1\pp$ (Section~\ref{sec:cost}). Teacher traces, where audited, are cleaner. Boundary cases are summarized in Table~\ref{tab:boundary}.}
\label{fig:cost}
\end{figure*}

Because LLM-judge magnitudes are prompt- and model-dependent \citep{llmjudgesurvey,ye2025justice}, we use the judge panel for direction and range rather than a single absolute score. Every retained configuration gives $\sft{>}\van$, with pooled effect sizes from $+7.7$ to $+19.8\pp$ and no reversal. The style-blind GLM prompt reduces the naive gap ($+7.7$ vs.\ $+14.3\pp$), but most of the inflation survives the instruction to ignore phrasing confidence (counts in App.~\ref{app:judges}). Figure~\ref{fig:cost} then shows the medical positive evidence, leaving boundary cases to Section~\ref{sec:boundary}.

The reliability checks target the main judge-artifact concern. GLM-4-32B and Hunyuan-A13B reach only moderate agreement on shared steps ($\kappa=0.434$), so their absolute labels are not interchangeable; even restricting to steps where they agree leaves a $+9.8\pp$ inflation. Abstention also does not carry the result. Smaller judges abstain often but roughly symmetrically, and Kimi gaps remain $+18.1$ and $+18.0\pp$ when every \texttt{uncertain} step is counted as correct or erroneous (App.~\ref{app:judges}).

Revealing the answer key to the judge is a second possible confound. Under an answer-blind Kimi prompt on the same 500-question subset, the gap remains: pooled error over non-\texttt{uncertain} steps is $28.8\%\!\to\!46.2\%$. Averaging the distilled-minus-base gap within each paired question gives $+14.2\pp$ ($95\%$ CI $[+11.8,+16.6]\pp$, Wilcoxon $p=4.2{\times}10^{-26}$; App.~\ref{app:controls}). Answer visibility is therefore not necessary for the overall reasoning-error increase.

The gap also survives the main accounting checks. Distilled chains are shorter, not longer (mean $7.0$ vs.\ $8.8$ steps); matching on step length leaves $+17.4\pp$; every step-position bin is positive ($+10.0$ to $+29.3\pp$); and excluding structural ``other'' steps still leaves $+17.0\pp$. Because the base model more often uses numbered scaffolding while the distilled model writes smoother prose, we also run direct surface controls on the first 500 MedQA questions. Tone-swap rewriting leaves a $+18.6\pp$ gap ($95\%$ CI $[+16.2,+21.0]\pp$), and common step-boundary rules remain positive: $+13.4\pp$ for sentence chunks and $+15.2\pp$ for fixed word windows (App.~\ref{app:controls}). These controls do not make LLM judging ground truth, but they close the most direct style and segmentation routes.

The result is not tied to one sampled chain. Since the main audit uses the first SC@64 sample, we re-audit chains 2--4 on a 200-question subset. Their average pooled gap is $+19.1\pp$, within $0.7\pp$ of the pooled full-set gap ($+19.8\pp$).

Two medical transfer checks keep the phenomenon inside clinical QA rather than one benchmark. On MedMCQA, the MedQA-trained distilled model is no more accurate than the base model ($59.4\%$ vs.\ $59.6\%$ greedy), yet step-error rate still rises from $32.2\%$ to $49.5\%$ ($+17.3\pp$). MedBullets5 tests harder clinical vignettes with an explicitly stronger teacher: on $298$ unique public questions, the teacher answers $76.5\%$ correctly and has the cleanest audited trace ($44.3\%$ segment error), while the distilled student improves over the base in accuracy ($53.4{\to}57.4\%$) but worsens in segment error ($65.5{\to}78.4\%$; paired increase $+10.0\pp$, $95\%$ CI $[+6.7,+13.2]\pp$). Details are in App.~\ref{app:medbullets}. The nonmedical audits play a different role: GSM8K, MATH, and ARC-Challenge mark where the medical effect stops.

Human review gives an independent check. Two clinical experts independently labelled the same 450 blinded steps (150 per model). Both preserve the judge panel's ordering, with post-stratified distilled-minus-base gaps of $+13.0\pp$ and $+12.9\pp$; teacher traces remain cleanest. Their agreement is $81.6\%$ (Cohen's $\kappa=0.556$). Most flagged errors involve substantive medical falsehoods (App.~\ref{app:judges}).

\section{Decomposing Accuracy Gains}
\label{sec:answer_gain}

The split is easier to interpret once we account for the answer gain itself. If the distilled trace is less reliable, how can SC@64 accuracy improve so much? The answer is not that distillation suddenly solves questions the base model could not touch. It mostly changes the winner among options the base model was already sampling.

For each MedQA question, we rank the gold option by its frequency in the base model's $64$ sampled answers. This rank is a pre-distillation reach proxy: rank $1$ means the base SC vote already selects the gold option, rank $2$ means the gold option is the runner-up, and ``absent'' means no base sample produced it. We also track $p_g$, the fraction of the $64$ samples that choose the gold option. Across all $1{,}273$ questions, mean $p_g$ rises from $71.4\%$ to $76.2\%$ after distillation ($+4.7\pp$, paired Wilcoxon $p=1.3{\times}10^{-10}$). The aggregate hides a sharper pattern. Distillation creates $159$ SC rescues and $40$ breaks, and most rescues come from near-frontier questions where the correct option was already the base model's second choice.

The gain has a more specific shape. Already-won rank-1 questions slightly lose gold vote share and account for the breaks. Fully absent questions gain some probability mass but rarely flip the SC decision. Most of the movement is at the decision boundary: the correct answer is already reachable, and distillation moves it from runner-up to winner.

The trace audit cuts through the same strata. The paired per-question step-error increase is positive inside every reach bucket (Table~\ref{tab:answer_gain_main}). The rank-2 bucket, which supplies most answer rescues, still worsens by $+19.3\pp$ ($95\%$ CI $[+15.3,+23.3]\pp$). Rank-1 questions, where the base answer was already correct, worsen by $+15.1\pp$ ($95\%$ CI $[+13.2,+17.0]\pp$). Pooling the lowest-reach buckets (rank $4$ or absent) gives $+17.2\pp$ ($95\%$ CI $[+10.1,+24.2]\pp$). Thus the trace degradation is not an artifact of newly rescued hard questions entering the audit. It is present within the same answer-reach regions where the answer policy improves.

This gives a concrete account of how the two metrics separate. SC@64 is a low-bandwidth contest over four options, where a modest redistribution of answer mass can change the winner; the trace is high-bandwidth, full of local claims the option letter never checks. Supervised fine-tuning can teach a competent student to put more probability on the right option near the vote boundary while also teaching it to write a teacher-like route whose local factual claims it does not reliably support. The answer becomes better selected; the reasoning text becomes less dependable.

\begin{table}[t]
\centering
{\scriptsize
\setlength{\tabcolsep}{2.5pt}
\begin{tabular}{@{}lrrrr@{}}
\toprule
Base reach & $N$ & $\Delta p_g$ & SC flips & $\Delta$ step err. \\
\midrule
rank $1$ & 951 & $-2.2$ & $0/40$ & $+15.1$ \\
rank $2$ & 196 & $+28.0$ & $123/0$ & $+19.3$ \\
rank $3$ & 56 & $+20.2$ & $21/0$ & $+20.4$ \\
rank $4$ & 22 & $+27.5$ & $8/0$ & $+23.0$ \\
absent & 48 & $+18.5$ & $7/0$ & $+14.0$ \\
\bottomrule
\end{tabular}
}
\caption{Answer gain by the gold option's rank in the base SC@64 distribution. $\Delta p_g$ is distilled-minus-base gold vote share. SC flips are rescues/breaks. $\Delta$ step error is the paired per-question Kimi-K2.6 gap within each bucket.}
\label{tab:answer_gain_main}
\end{table}

\section{The Medical Risk Regime}
\label{sec:boundary}

The positive evidence in this paper is medical because the mechanism is clearest where labels compress rich clinical rationales. The relevant axis is process--answer coupling: how strongly the signal that says ``this answer is good'' constrains the local claims in the emitted route. Teacher step cleanliness is a diagnostic of that coupling, not the cause itself. A high-accuracy teacher can still be a poor process target if its derivation is compressed, lossy, or locally unreliable; a clean teacher can still yield a noisier student when a capable but imperfect student learns answer-selection and discourse patterns without the knowledge needed to instantiate every local claim. This is the trace-level version of a broader lesson from ``right for the right reasons'' and shortcut-learning work \citep{ross2017right,geirhos2020shortcut}.

Medical diagnosis sits near the risky end of this axis. A label such as AD, MCI, or cognitively normal is a compact endpoint, while the rationale ranges over symptoms, functional decline, competing causes, and tests whose interpretation is not uniquely specified by the label. The same structure appears in exam-style clinical QA: the option key can verify which diagnosis was selected, but it cannot verify all the local medical claims used to get there. That combination makes teacher traces attractive supervision in exactly the domain where human step-level review is costly and where the trace is often the part a reader most wants.

The boundary checks ask which part of that structure fails when the medical pattern does not appear.

\begin{table}[t]
\centering
{\scriptsize
\setlength{\tabcolsep}{2.5pt}
\renewcommand{\arraystretch}{1.15}
\begin{tabular}{@{}L{0.22\columnwidth}L{0.31\columnwidth}L{0.39\columnwidth}@{}}
\toprule
Boundary check & Boundary cue & Observed pattern \\
\midrule
GSM8K, MATH & Teacher answer strength does not imply cleaner steps & GSM8K step error is 6.7/8.8/11.3\% for base/teacher/distilled; MATH shows the same teacher/base mismatch. \\
ARC-Challenge & The rationale is short and tightly tied to the answer & Accuracy is near ceiling ($92.6{\to}94.6$), and pooled step error does not rise ($9.2{\to}8.6$). \\
Mistral-7B on MedQA & The base student is below the risky imitation regime & SFT improves both accuracy ($48.6{\to}67.6$) and step error ($56.5{\to}49.1$). \\
\bottomrule
\end{tabular}
}
\caption{Boundary diagnostics for the medical-risk regime. These negative controls weaken different ingredients of the positive medical setting, so they mark where the accuracy--reasoning split should not be expected.}
\label{tab:boundary}
\end{table}

Read together, these boundary cases delimit the medical result. The risk is not supervised fine-tuning itself. It comes from medical-rationale distillation under weak answer-to-process supervision: a compact answer supervises a rich clinical rationale, and the student is capable enough to imitate expert-like form without reliably grounding each local claim.

\section{Factual Errors Across Reasoning Roles}
\label{sec:taxonomy}

If the loss were merely a style artifact, it should collect in a narrow corner of the trace. Instead, the degradation appears within every reasoning-bearing role. We classify steps as hypothesis generation / differential diagnosis, option elimination, correction / backtracking, factual claim, final synthesis, or structural other, and recompute the error rate inside each role, excluding ``other''.

The taxonomy is intentionally functional rather than linguistic. A step is hypothesis-generation when it opens or weighs a diagnostic possibility, option-elimination when it rules an answer in or out, correction when it revises an earlier claim, and factual claim when it states a medical mechanism, value, or association without doing one of the more specific reasoning acts (classifier details in App.~\ref{app:taxonomy}). This matters because the main question is not whether distillation changes surface form. It is whether comparable pieces of reasoning become less dependable after the student has been trained to imitate a stronger teacher's CoT style.

\begin{figure}[t]
\centering
\includegraphics[width=\columnwidth]{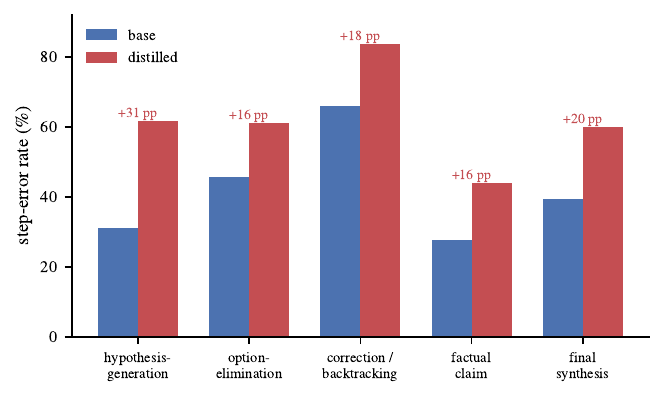}
\caption{Step-level factual-error rates by reasoning role before and after distillation, using the primary judge with access to the answer key. Results cover the full test set and exclude structural ``other'' steps. The answer-blind control is reported in App.~\ref{app:controls}.}
\label{fig:taxonomy}
\end{figure}

The largest degradation appears in the most diagnosis-sensitive role. Hypothesis-generation and differential-diagnosis steps move from $31.1\%$ error to $61.8\%$, a $+30.7\pp$ jump that roughly doubles the within-role baseline rate. The other reasoning-bearing roles also degrade, but cluster in the narrower $+15.6$ to $+20.4\pp$ band (Figure~\ref{fig:taxonomy}). The trace therefore becomes least reliable at the point where a reader most needs the model to be careful: when it is opening or weighing the diagnostic space.

This role-wise peak is a judged factuality result, not a claim that every non-gold hypothesis is intrinsically false. The answer-blind control in App.~\ref{app:controls} keeps the overall gap large while reducing the special leverage of the gold answer, so we use the taxonomy to locate flagged errors and the answer-blind audit to support the overall degradation.

This is not simply ``more search''. The distilled model's exploratory share is slightly lower ($23.8\%$ vs.\ $26.0\%$), and the count of erroneous exploratory steps per chain is essentially unchanged. What changes is reliability given the role. Many degraded steps remain locally plausible: they name the right disease class or point toward the correct option, but distort a differential, overstate an association, or smuggle in a false exclusion.

This role-wise pattern explains why answer filtering does not solve the problem. A trace can contain a false exclusion and still land on the right option because later steps recover, the wrong claim concerns a distractor, or self-consistency voting favours the correct answer despite a noisy rationale. The answer-conditioned analysis in App.~\ref{app:answer_conditioned} is the empirical version: even when both audited chains end correctly, the distilled rationale remains substantially noisier.

\section{Standard Signals and Simple Alternatives}
\label{sec:invisible}
\label{sec:notrat}
\label{sec:fix}

The surprising part is not only that trace factuality drops. It is that the usual signals stay calm. Accuracy, ECE, and Brier improve; aggregate hedging barely moves ($0.40{\to}0.42$); and within each trace, hedge words still mark shakier local claims. In both $\van$ and $\sft$, a hedged step is about $24\pp$ more likely to be judged erroneous than a non-hedged step ($+24.6\pp$ and $+23.1\pp$; point-biserial $r{=}0.30$ and $0.24$). What hedging misses is the global shift: the reader sees a familiar density of caution words, while the trace sits on a much higher error baseline. One might hope that the shift is at least available to a simple internal diagnostic. We therefore fit a linear probe to residual activations at step boundaries; it predicts judge-labelled errors at AUROC $\approx 0.82$ in both models (App.~\ref{app:probe}), but step position and surface correlates explain part of that power. The blind spot is therefore not the absence of every signal. It is that answer metrics, visible hedging, and an internal diagnostic all fail to carry the new reliability baseline outward.

Two deflationary explanations would make the shift less informative. The first is answer-first rationalization: the distilled student commits to an answer, then writes claims that make it look supported. We test this by appending an answer prompt to successive chain prefixes and reading the model's A/B/C/D logits (App.~\ref{app:rationalization}). The direction is the opposite of the prediction: pre-CoT answer probability falls ($0.75{\to}0.64$), lock step moves later ($1.35{\to}1.69$), the mid-CoT minimum drops ($0.69{\to}0.49$), and explicit backtracking becomes more frequent ($6.8\%{\to}8.2\%$ of steps). This rules out the simple account in which the trace is only a decorative after-the-fact rationale.

The second is a broken caution channel. A targeted activation intervention rules it out. In the residual stream (the hidden activation vector passed between transformer blocks), we find a one-dimensional late-layer direction whose removal nearly eliminates hedge-token probability; on a held-out set, projection-out reduces next-token probability mass on a fixed hedge-marker list from $0.504$ to $0.0017$ ($99.7\%$). A separate patching analysis finds recovery peaks in the last $\sim$10\% of layers across Qwen3-8B, Llama-3.1-8B, and Mistral-7B (App.~\ref{app:hedge}). This check is not a mitigation or a full cause of the factuality drop. It shows that the caution channel remains available after distillation but is no longer calibrated to the new reliability level.

\section{Discussion and Conclusion}\label{sec:discussion}

Answer metrics are the right metrics when the final output is the only product. A medical reasoning trace is different: it is read as a justification, monitored as a safety surface, and reused as training data. Our results show that these two products can move in opposite directions. The answer improves, while the written process becomes a worse object to read or reuse.

The regime is conditional but important. The risk appears when answer-level supervision is weakly coupled to trace-level correctness, a capable student has answer headroom, and the trace is a high-bandwidth object whose local claims are only weakly checked by the final answer. That describes diagnostic medical rationales, where compact labels compress symptoms, tests, exclusions, and competing diagnoses. A distilled medical reasoning model that exposes traces should therefore report a small process audit alongside answer metrics.

The training-data problem is the sharpest implication. CoT distillation is often recursive: one model's traces become another model's supervision, and answer-filtered traces are treated as a cheap substitute for step-level annotation. Our answer-conditioned analysis shows why that is risky: even when both audited chains end correctly, the distilled rationale is substantially noisier. Medical CoT distillation has two products, not one. If the trace is shown to users, used for oversight, or fed back into training, it needs its own quality control.

\section*{Limitations}

The step labels primarily come from LLM judges. Our panel, style-blind prompt, agreement checks, tone-swap, answer-blind and step-boundary controls, and 450-step audit independently labelled by two clinical experts support the direction of the effect, not a calibrated prevalence estimate. Absolute rates remain judge-dependent, and the human audit is limited to a stratified sample of five reasoning roles. Revealing the answer key makes isolated steps more interpretable but can sharpen the role-wise peak; the answer-blind control preserves the overall gap but does not replace expert adjudication.

The audit measures one property of trace quality: factual reliability of emitted steps. It does not score completeness, pedagogy, causal faithfulness to the model's computation, or whether the trace helped the model arrive at the answer. The scope is medical rather than universal: MedQA, MedMCQA, and MedBullets5 are positive replications, while GSM8K, MATH, ARC-Challenge, and weak-student Mistral are boundary diagnostics. Student-side checks vary scale and family rather than the teacher family. The mechanism remains incomplete: our answer-first and hedge-channel checks rule out two simple accounts, but they do not identify a full training-time cause or mitigation.

All datasets and models used are publicly available for research and are used consistent with their intended research use.

\bibliography{refs}

@article{deepseekr1,
  title={Deepseek-r1: Incentivizing reasoning capability in llms via reinforcement learning},
  author={Guo, Daya and Yang, Dejian and Zhang, Haowei and Song, Junxiao and Wang, Peiyi and Zhu, Qihao and Xu, Runxin and Zhang, Ruoyu and Ma, Shirong and Bi, Xiao and others},
  journal={arXiv preprint arXiv:2501.12948},
  year={2025}
}

@inproceedings{magister2023teaching,
  title={Teaching small language models to reason},
  author={Magister, Lucie Charlotte and Mallinson, Jonathan and Adamek, Jakub and Malmi, Eric and Severyn, Aliaksei},
  booktitle={Proceedings of the 61st Annual Meeting of the Association for Computational Linguistics (Volume 2: Short Papers)},
  pages={1773--1781},
  year={2023}
}

@inproceedings{hsieh2023distilling,
  title={Distilling step-by-step! outperforming larger language models with less training data and smaller model sizes},
  author={Hsieh, Cheng-Yu and Li, Chun-Liang and Yeh, Chih-Kuan and Nakhost, Hootan and Fujii, Yasuhisa and Ratner, Alex and Krishna, Ranjay and Lee, Chen-Yu and Pfister, Tomas},
  booktitle={Findings of the Association for Computational Linguistics: ACL 2023},
  pages={8003--8017},
  year={2023}
}

@inproceedings{ho2023large,
  title={Large language models are reasoning teachers},
  author={Ho, Namgyu and Schmid, Laura and Yun, Se-Young},
  booktitle={Proceedings of the 61st annual meeting of the association for computational linguistics (volume 1: long papers)},
  pages={14852--14882},
  year={2023}
}

@inproceedings{lightman2024lets,
  title={Let's verify step by step},
  author={Lightman, Hunter and Kosaraju, Vineet and Burda, Yuri and Edwards, Harrison and Baker, Bowen and Lee, Teddy and Leike, Jan and Schulman, John and Sutskever, Ilya and Cobbe, Karl},
  booktitle={International Conference on Learning Representations},
  volume={2024},
  pages={39578--39601},
  year={2024}
}

@article{uesato2022solving,
  title={Solving math word problems with process-and outcome-based feedback},
  author={Uesato, Jonathan and Kushman, Nate and Kumar, Ramana and Song, Francis and Siegel, Noah and Wang, Lisa and Creswell, Antonia and Irving, Geoffrey and Higgins, Irina},
  journal={arXiv preprint arXiv:2211.14275},
  year={2022}
}

@article{turpin2023language,
  title={Language models don't always say what they think: Unfaithful explanations in chain-of-thought prompting},
  author={Turpin, Miles and Michael, Julian and Perez, Ethan and Bowman, Samuel},
  journal={Advances in Neural Information Processing Systems},
  volume={36},
  pages={74952--74965},
  year={2023}
}

@article{lanham2023measuring,
  title={Measuring faithfulness in chain-of-thought reasoning},
  author={Lanham, Tamera and Chen, Anna and Radhakrishnan, Ansh and Steiner, Benoit and Denison, Carson and Hernandez, Danny and Li, Dustin and Durmus, Esin and Hubinger, Evan and Kernion, Jackson and others},
  journal={arXiv preprint arXiv:2307.13702},
  year={2023}
}

@article{baker2025monitoring,
  title={Monitoring reasoning models for misbehavior and the risks of promoting obfuscation},
  author={Baker, Bowen and Huizinga, Joost and Gao, Leo and Dou, Zehao and Guan, Melody Y and Madry, Aleksander and Zaremba, Wojciech and Pachocki, Jakub and Farhi, David},
  journal={arXiv preprint arXiv:2503.11926},
  year={2025}
}

@article{korbak2025chainofthought,
  title={Chain of thought monitorability: A new and fragile opportunity for ai safety},
  author={Korbak, Tomek and Balesni, Mikita and Barnes, Elizabeth and Bengio, Yoshua and Benton, Joe and Bloom, Joseph and Chen, Mark and Cooney, Alan and Dafoe, Allan and Dragan, Anca and others},
  journal={arXiv preprint arXiv:2507.11473},
  year={2025}
}

@article{lin2022teaching,
  title={Teaching models to express their uncertainty in words},
  author={Lin, Stephanie and Hilton, Jacob and Evans, Owain},
  journal={arXiv preprint arXiv:2205.14334},
  year={2022}
}

@article{kadavath2022language,
  title={Language models (mostly) know what they know},
  author={Kadavath, Saurav and Conerly, Tom and Askell, Amanda and Henighan, Tom and Drain, Dawn and Perez, Ethan and Schiefer, Nicholas and Hatfield-Dodds, Zac and DasSarma, Nova and Tran-Johnson, Eli and others},
  journal={arXiv preprint arXiv:2207.05221},
  year={2022}
}

@article{jin2021disease,
  title={What disease does this patient have? a large-scale open domain question answering dataset from medical exams},
  author={Jin, Di and Pan, Eileen and Oufattole, Nassim and Weng, Wei-Hung and Fang, Hanyi and Szolovits, Peter},
  journal={Applied Sciences},
  volume={11},
  number={14},
  pages={6421},
  year={2021},
  publisher={MDPI}
}

@article{arditi2024refusal,
  title={Refusal in language models is mediated by a single direction},
  author={Arditi, Andy and Obeso, Oscar and Syed, Aaquib and Paleka, Daniel and Panickssery, Nina and Gurnee, Wes and Nanda, Neel},
  journal={Advances in Neural Information Processing Systems},
  volume={37},
  pages={136037--136083},
  year={2024}
}

@inproceedings{zhang2024towards,
  title={Towards best practices of activation patching in language models: Metrics and methods},
  author={Zhang, Fred and Nanda, Neel},
  booktitle={International Conference on Learning Representations},
  volume={2024},
  pages={1651--1678},
  year={2024}
}

@inproceedings{elhage2021mathematical,
 author = {Meng, Kevin and Bau, David and Andonian, Alex and Belinkov, Yonatan},
 booktitle = {Advances in Neural Information Processing Systems},
 doi = {10.52202/068431-1262},
 editor = {S. Koyejo and S. Mohamed and A. Agarwal and D. Belgrave and K. Cho and A. Oh},
 pages = {17359--17372},
 publisher = {Curran Associates, Inc.},
 title = {Locating and Editing Factual Associations in GPT},
 url = {https://proceedings.neurips.cc/paper_files/paper/2022/file/6f1d43d5a82a37e89b0665b33bf3a182-Paper-Conference.pdf},
 volume = {35},
 year = {2022}
}

@inproceedings{wadhwa2024investigating,
  title={Investigating mysteries of cot-augmented distillation},
  author={Wadhwa, Somin and Amir, Silvio and Wallace, Byron C},
  booktitle={Proceedings of the 2024 Conference on Empirical Methods in Natural Language Processing},
  pages={6071--6086},
  year={2024}
}

@inproceedings{ranaldi2024aligning,
  title={Aligning large and small language models via chain-of-thought reasoning},
  author={Ranaldi, Leonardo and Freitas, Andre},
  booktitle={Proceedings of the 18th Conference of the European Chapter of the Association for Computational Linguistics (Volume 1: Long Papers)},
  pages={1812--1827},
  year={2024}
}

@inproceedings{sun2025reasonmed,
    title = "{R}eason{M}ed: A 370{K} Multi-Agent Generated Dataset for Advancing Medical Reasoning",
    author = "Sun, Yu  and
      Qian, Xingyu  and
      Xu, Weiwen  and
      Zhang, Hao  and
      Xiao, Chenghao  and
      Li, Long  and
      Zhao, Deli  and
      Huang, Wenbing  and
      Xu, Tingyang  and
      Bai, Qifeng  and
      Rong, Yu",
    editor = "Christodoulopoulos, Christos  and
      Chakraborty, Tanmoy  and
      Rose, Carolyn  and
      Peng, Violet",
    booktitle = "Proceedings of the 2025 Conference on Empirical Methods in Natural Language Processing",
    month = nov,
    year = "2025",
    address = "Suzhou, China",
    publisher = "Association for Computational Linguistics",
    url = "https://aclanthology.org/2025.emnlp-main.1344/",
    doi = "10.18653/v1/2025.emnlp-main.1344",
    pages = "26446--26467",
    ISBN = "979-8-89176-332-6",
}

@article{wu2025medreason,
  title={Medreason: Eliciting factual medical reasoning steps in llms via knowledge graphs},
  author={Wu, Juncheng and Deng, Wenlong and Li, Xingxuan and Liu, Sheng and Mi, Taomian and Peng, Yifan and Xu, Ziyang and Liu, Yi and Cho, Hyunjin and Choi, Chang-In and others},
  journal={arXiv preprint arXiv:2504.00993},
  year={2025}
}

@inproceedings{
hu2022lora,
title={Lo{RA}: Low-Rank Adaptation of Large Language Models},
author={Edward J Hu and Yelong Shen and Phillip Wallis and Zeyuan Allen-Zhu and Yuanzhi Li and Shean Wang and Lu Wang and Weizhu Chen},
booktitle={International Conference on Learning Representations},
year={2022},
url={https://openreview.net/forum?id=nZeVKeeFYf9}
}

@article{qwen3,
  title={Qwen3 technical report},
  author={Yang, An and Li, Anfeng and Yang, Baosong and Zhang, Beichen and Hui, Binyuan and Zheng, Bo and Yu, Bowen and Gao, Chang and Huang, Chengen and Lv, Chenxu and others},
  journal={arXiv preprint arXiv:2505.09388},
  year={2025}
}

@article{deepseekv3,
  title={Deepseek-v3 technical report},
  author={Liu, Aixin and Feng, Bei and Xue, Bing and Wang, Bingxuan and Wu, Bochao and Lu, Chengda and Zhao, Chenggang and Deng, Chengqi and Zhang, Chenyu and Ruan, Chong and others},
  journal={arXiv preprint arXiv:2412.19437},
  year={2024}
}

@inproceedings{azaria2023internal,
  title={The internal state of an LLM knows when it’s lying},
  author={Azaria, Amos and Mitchell, Tom},
  booktitle={Findings of the Association for Computational Linguistics: EMNLP 2023},
  pages={967--976},
  year={2023}
}

@article{burns2022discovering,
  title={Discovering latent knowledge in language models without supervision},
  author={Burns, Collin and Ye, Haotian and Klein, Dan and Steinhardt, Jacob},
  journal={arXiv preprint arXiv:2212.03827},
  year={2022}
}

@inproceedings{tian2023just,
  title={Just ask for calibration: Strategies for eliciting calibrated confidence scores from language models fine-tuned with human feedback},
  author={Tian, Katherine and Mitchell, Eric and Zhou, Allan and Sharma, Archit and Rafailov, Rafael and Yao, Huaxiu and Finn, Chelsea and Manning, Christopher D},
  booktitle={Proceedings of the 2023 Conference on Empirical Methods in Natural Language Processing},
  pages={5433--5442},
  year={2023}
}

@inproceedings{zhou2024relying,
  title={Relying on the unreliable: The impact of language models’ reluctance to express uncertainty},
  author={Zhou, Kaitlyn and Hwang, Jena and Ren, Xiang and Sap, Maarten},
  booktitle={Proceedings of the 62nd Annual Meeting of the Association for Computational Linguistics (Volume 1: Long Papers)},
  pages={3623--3643},
  year={2024}
}

@article{mielke2022reducing,
  title={Reducing conversational agents’ overconfidence through linguistic calibration},
  author={Mielke, Sabrina J and Szlam, Arthur and Dinan, Emily and Boureau, Y-Lan},
  journal={Transactions of the Association for Computational Linguistics},
  volume={10},
  pages={857--872},
  year={2022},
  publisher={MIT Press One Broadway, 12th Floor, Cambridge, Massachusetts 02142, USA~…}
}

@inproceedings{paul2024making,
  title={Making reasoning matter: Measuring and improving faithfulness of chain-of-thought reasoning},
  author={Paul, Debjit and West, Robert and Bosselut, Antoine and Faltings, Boi},
  booktitle={Findings of the Association for Computational Linguistics: EMNLP 2024},
  pages={15012--15032},
  year={2024}
}

@article{stechly2024chain,
  title={Chain of thoughtlessness? an analysis of cot in planning},
  author={Stechly, Kaya and Valmeekam, Karthik and Kambhampati, Subbarao},
  journal={Advances in Neural Information Processing Systems},
  volume={37},
  pages={29106--29141},
  year={2024}
}

@inproceedings{pal2022medmcqa,
  title={Medmcqa: A large-scale multi-subject multi-choice dataset for medical domain question answering},
  author={Pal, Ankit and Umapathi, Logesh Kumar and Sankarasubbu, Malaikannan},
  booktitle={Conference on health, inference, and learning},
  pages={248--260},
  year={2022},
  organization={PMLR}
}

@article{zheng2023judging,
  title={Judging llm-as-a-judge with mt-bench and chatbot arena},
  author={Zheng, Lianmin and Chiang, Wei-Lin and Sheng, Ying and Zhuang, Siyuan and Wu, Zhanghao and Zhuang, Yonghao and Lin, Zi and Li, Zhuohan and Li, Dacheng and Xing, Eric and others},
  journal={Advances in neural information processing systems},
  volume={36},
  pages={46595--46623},
  year={2023}
}

@inproceedings{wu2025style,
  title={Style over substance: Evaluation biases for large language models},
  author={Wu, Minghao and Aji, Alham Fikri},
  booktitle={Proceedings of the 31st International Conference on Computational Linguistics},
  pages={297--312},
  year={2025}
}

@article{llmjudgesurvey,
title = {A survey on LLM-as-a-judge},
journal = {The Innovation},
volume = {7},
number = {6},
pages = {101253},
year = {2026},
issn = {2666-6758},
doi = {https://doi.org/10.1016/j.xinn.2025.101253},
url = {https://www.sciencedirect.com/science/article/pii/S2666675825004564},
author = {Jiawei Gu and Xuhui Jiang and Zhichao Shi and Hexiang Tan and Xuehao Zhai and Chengjin Xu and Wei Li and Yinghan Shen and Shengjie Ma and Honghao Liu and Saizhuo Wang and Kun Zhang and Zhouchi Lin and Bowen Zhang and Lionel Ni and Wen Gao and Yuanzhuo Wang and Jian Guo},
}

@inproceedings{ye2025justice,
  title={Justice or prejudice? quantifying biases in llm-as-a-judge},
  author={Ye, Jiayi and Wang, Yanbo and Huang, Yue and Chen, Dongping and Zhang, Qihui and Moniz, Nuno and Gao, Tian and Geyer, Werner and Huang, Chao and Chen, Pin-Yu and others},
  booktitle={International Conference on Learning Representations},
  volume={2025},
  pages={102351--102390},
  year={2025}
}

@article{wu2025whycot,
  title={Why chain of thought fails in clinical text understanding},
  author={Wu, Jiageng and Xie, Kevin and Gu, Bowen and Kr{\"u}ger, Nils and Lin, Kueiyu Joshua and Yang, Jie},
  journal={arXiv preprint arXiv:2509.21933},
  year={2025}
}

@article{yao2025reasoning,
  title={Are reasoning models more prone to hallucination?},
  author={Yao, Zijun and Liu, Yantao and Chen, Yanxu and Chen, Jianhui and Fang, Junfeng and Hou, Lei and Li, Juanzi and Chua, Tat-Seng},
  journal={arXiv preprint arXiv:2505.23646},
  year={2025}
}

@article{cobbe2021training,
  title={Training verifiers to solve math word problems},
  author={Cobbe, Karl and Kosaraju, Vineet and Bavarian, Mohammad and Chen, Mark and Jun, Heewoo and Kaiser, Lukasz and Plappert, Matthias and Tworek, Jerry and Hilton, Jacob and Nakano, Reiichiro and others},
  journal={arXiv preprint arXiv:2110.14168},
  year={2021}
}

@article{petersen1999mild,
  title={Mild cognitive impairment: clinical characterization and outcome},
  author={Petersen, Ronald C and Smith, Glenn E and Waring, Stephen C and Ivnik, Robert J and Tangalos, Eric G and Kokmen, Emre},
  journal={Archives of neurology},
  volume={56},
  number={3},
  pages={303--308},
  year={1999},
  publisher={American Medical Association}
}

@article{albert2011diagnosis,
  title={The diagnosis of mild cognitive impairment due to Alzheimer's disease: recommendations from the National Institute on Aging-Alzheimer's Association workgroups on diagnostic guidelines for Alzheimer's disease},
  author={Albert, Marilyn S and DeKosky, Steven T and Dickson, Dennis and Dubois, Bruno and Feldman, Howard H and Fox, Nick C and Gamst, Anthony and Holtzman, David M and Jagust, William J and Petersen, Ronald C and others},
  journal={Alzheimer's \& dementia},
  volume={7},
  number={3},
  pages={270--279},
  year={2011},
  publisher={Wiley Online Library}
}

@article{mckhann2011diagnosis,
  title={The diagnosis of dementia due to Alzheimer's disease: recommendations from the National Institute on Aging-Alzheimer's Association workgroups on diagnostic guidelines for Alzheimer's disease},
  author={McKhann, Guy M and Knopman, David S and Chertkow, Howard and Hyman, Bradley T and Jack Jr, Clifford R and Kawas, Claudia H and Klunk, William E and Koroshetz, Walter J and Manly, Jennifer J and Mayeux, Richard and others},
  journal={Alzheimer's \& dementia},
  volume={7},
  number={3},
  pages={263--269},
  year={2011},
  publisher={Wiley Online Library}
}

@inproceedings{chen-etal-2025-benchmarking,
  title={Benchmarking large language models on answering and explaining challenging medical questions},
  author={Chen, Hanjie and Fang, Zhouxiang and Singla, Yash and Dredze, Mark},
  booktitle={Proceedings of the 2025 Conference of the Nations of the Americas Chapter of the Association for Computational Linguistics: Human Language Technologies (Volume 1: Long Papers)},
  pages={3563--3599},
  year={2025}
}

@article{hendrycks2021math,
  title={Measuring mathematical problem solving with the math dataset},
  author={Hendrycks, Dan and Burns, Collin and Kadavath, Saurav and Arora, Akul and Basart, Steven and Tang, Eric and Song, Dawn and Steinhardt, Jacob},
  journal={arXiv preprint arXiv:2103.03874},
  year={2021}
}

@article{clark2018think,
  title={Think you have solved question answering? try arc, the ai2 reasoning challenge},
  author={Clark, Peter and Cowhey, Isaac and Etzioni, Oren and Khot, Tushar and Sabharwal, Ashish and Schoenick, Carissa and Tafjord, Oyvind},
  journal={arXiv preprint arXiv:1803.05457},
  year={2018}
}

@article{ross2017right,
  title={Right for the right reasons: Training differentiable models by constraining their explanations},
  author={Ross, Andrew Slavin and Hughes, Michael C and Doshi-Velez, Finale},
  journal={arXiv preprint arXiv:1703.03717},
  year={2017}
}

@article{geirhos2020shortcut,
  title={Shortcut learning in deep neural networks},
  author={Geirhos, Robert and Jacobsen, J{\"o}rn-Henrik and Michaelis, Claudio and Zemel, Richard and Brendel, Wieland and Bethge, Matthias and Wichmann, Felix A},
  journal={Nature Machine Intelligence},
  volume={2},
  number={11},
  pages={665--673},
  year={2020},
  publisher={Nature Publishing Group UK London}
}

@article{bassil2023feasibility,
  title={Feasibility of an online consensus approach for the diagnosis of cognitive impairment and dementia in rural South Africa},
  author={Bassil, Darina T and Farrell, Meagan T and Weerman, Albert and Guo, Muqi and Wagner, Ryan G and Brickman, Adam M and Glymour, M Maria and Langa, Kenneth M and Manly, Jennifer J and Tipping, Brent and others},
  journal={Alzheimer's \& Dementia: Diagnosis, Assessment \& Disease Monitoring},
  volume={15},
  number={2},
  pages={e12420},
  year={2023},
  publisher={Wiley Online Library}
}

@article{jiang2026lov3d,
  title={{LoV3D}: Grounding Cognitive Prognosis Reasoning in Longitudinal {3D} Brain {MRI} via Regional Volume Assessments},
  author={Jiang, Zhaoyang and Fu, Zhizhong and McAllister, David and Kim, Yunsoo and Wu, Honghan},
  journal={arXiv preprint arXiv:2603.12071},
  year={2026}
}
\appendix

\section{Step-Level Audit Protocol}
\label{app:audit}
\paragraph{Style-blind judge prompt.} Each step is judged in isolation with the following template (the question, the four options, the answer key, and the single step are substituted in):
\begin{quote}\small\ttfamily
You are a senior physician evaluating a medical AI model's reasoning. [question, options, correct answer, one reasoning step] \ldots You MUST evaluate ONLY whether the FACTUAL CONTENT is wrong. IGNORE: the tone or register (committal vs.\ hedged: ``X is'' vs.\ ``X may be''); grammatical confidence; whether the writer used hedging words; stylistic certainty markers; whether the step sounds confident or uncertain. ONLY judge: ``If I IGNORE the tone and read the underlying claim, is the medical fact WRONG?'' [four worked examples] Respond exactly: JUDGMENT: <correct$|$error$|$uncertain>; EXPLANATION: <one sentence>.
\end{quote}
The non-style-blind variant drops the IGNORE block and the examples. The full prompts are in the released code.
\paragraph{Segmentation.} Each CoT is split at lines beginning with an enumerator---``\texttt{N.}'', ``\texttt{N)}'', ``\texttt{N:}'', or ``\texttt{Step N.}''---taking the text up to the next such enumerator as one step (capped at 12 steps, each truncated to 800 characters; markdown headers and items shorter than 20 characters are dropped). If no numbered list is found, paragraphs of $\ge$30 characters are used as steps. The exact regular expression is in the released code; the same rule is used for the probe (Appendix~\ref{app:probe}) and the taxonomy (Appendix~\ref{app:taxonomy}), so step indices align with the judge labels.
\paragraph{CoT source.} For each model we audit one CoT per question---the first of the 64 self-consistency samples (temperature $\approx 0.7$)---over the 1{,}273-question MedQA-USMLE test set for the primary Kimi-K2.6 audit, including the teacher traces. The other judge configurations use a random 500-question subset. The step-error rate is $n_{\mathrm{err}}/(n_{\mathrm{err}}+n_{\mathrm{corr}})$ over steps labelled \texttt{error} or \texttt{correct}; \texttt{uncertain} steps (the judge declined to call) are excluded from the denominator.

\section{Answer-conditioned trace reliability}
\label{app:answer_conditioned}
The answer-conditioned analysis uses the final option written in the same sampled chain that is audited, not the SC@64 vote. This makes the conditioning deliberately strict: it asks whether traces remain noisier even when the emitted answer in that trace is already correct. Among the $784$ questions where both $\van$ and $\sft$ answer correctly, $\van$ has $4879$ correct, $1386$ error, and $659$ \texttt{uncertain} step judgments ($22.1\%$ committed error), while $\sft$ has $2950$ correct, $1770$ error, and $504$ \texttt{uncertain} judgments ($37.5\%$). The paired per-question gap is $+12.1\pp$ with bootstrap $95\%$ CI $[+10.2,+14.1]\pp$ and Wilcoxon $p=1.9{\times}10^{-30}$ over $657$ paired questions.

The same direction holds under other answer strata. When both sampled chains are wrong ($207$ questions), error rates are $49.4\%\!\to\!74.8\%$ and the paired gap is $+24.2\pp$ ($95\%$ CI $[+20.6,+27.8]\pp$). Conditioning only on questions where the distilled chain is correct ($943$ questions) still gives $25.9\%\!\to\!40.5\%$, with paired gap $+11.0\pp$ ($[+9.2,+12.9]\pp$). The point is not that correct and incorrect final answers are equally reliable. It is that answer-level filtering does not remove the trace-level degradation.

\section{Answer-gain accounting details}
\label{app:answer_gain}
Section~\ref{sec:answer_gain} uses the base model's SC@64 answer distribution as a pre-distillation reach proxy. Options are ranked by their frequency among the $64$ base samples: rank $1$ means the gold option wins the vote, rank $2$ means it is the runner-up, and ``absent'' means it was never sampled. We report $p_{\mathrm{gold}}$, the fraction of samples selecting the gold option.

Across all $1{,}273$ MedQA questions, mean $p_{\mathrm{gold}}$ rises from $71.4\%$ to $76.2\%$ after distillation ($+4.7\pp$, paired Wilcoxon $p=1.3{\times}10^{-10}$). Table~\ref{tab:answer_gain_decomp} locates most rescues at rank $2$, near the decision boundary. Rank-1 questions account for the observed breaks, while fully absent questions are rarely rescued.

\begin{table}[h]
\centering
\scriptsize
\setlength{\tabcolsep}{2pt}
\begin{tabular}{@{}lrrrr@{}}
\toprule
Gold rank & $N$ & $\Delta p_g$ & Rescues & Breaks \\
\midrule
rank $1$ & 951 & $-2.2\,[-3.1,-1.3]$ & 0 & 40 \\
rank $2$ & 196 & $+28.0\,[+24.6,+31.3]$ & 123 & 0 \\
rank $3$ & 56 & $+20.2\,[+14.9,+25.9]$ & 21 & 0 \\
rank $4$ & 22 & $+27.5\,[+20.0,+35.9]$ & 8 & 0 \\
absent & 48 & $+18.5\,[+13.0,+24.8]$ & 7 & 0 \\
\bottomrule
\end{tabular}
\caption{Answer-gain decomposition by the gold option's rank in the base model's SC@64 answer distribution. $\Delta p_g$ is the distilled-minus-base change in the fraction of sampled answers that choose the gold option, with bootstrap $95\%$ CIs over questions. Rescues are questions where SC@64 changes from wrong to correct; breaks are correct to wrong.}
\label{tab:answer_gain_decomp}
\end{table}

The rank pattern is not just a post-hoc binning of MedQA. Using only features of the base model's answer distribution, a logistic model predicts whether a base-missed question will be rescued by the distilled model with five-fold MedQA AUROC $0.763$; a pass@$k$-only version gives AUROC $0.768$. Applying the same feature family out of distribution gives AUROC $0.786$ on MedMCQA and $0.813$ on MedBullets5. These classifiers are not part of the main claim, but they support the interpretation that rescues are legible from the base model's pre-distillation answer distribution.

The same frontier shape appears in the raw rescue counts. Among base-missed questions, rank-2 rescues are common on MedQA ($123/196$) and MedBullets5 ($22/47$), and remain the largest rescue source on MedMCQA ($225/712$ over all $4{,}183$ validation questions). Questions where the correct answer is absent from all $64$ base samples are rarely rescued: $7/48$ on MedQA, $4/287$ on MedMCQA, and $4/78$ on MedBullets5. The absolute rescue rates vary by benchmark, but the qualitative location of the gain is stable: distillation most often helps when the correct answer is already in the base model's sample support.

\section{Answer-first rationalization check}
\label{app:rationalization}
Section~\ref{sec:fix} tests whether the distilled model writes an after-the-fact rationale for an answer it has already chosen. A simple answer-first account makes three predictions. Before any rationale is written, the model should already assign high probability to the option it will eventually print. Once the rationale begins, the decoded option should lock early and remain stable. The probability of the eventual option should also avoid large mid-chain dips.

The anchor is the model's own final answer, not the gold label. For each MedQA test question and each model, we greedily generate a full chain, parse it into at most $12$ steps, and extract the final option printed by the model; call this option $\hat{y}$. The chain may be correct or incorrect. We then evaluate the empty-CoT prefix ($k=0$) and the same generated chain truncated after steps $1,\ldots,\min(n_{\mathrm{steps}},10)$. We stop at ten prefixes because the final parsed steps often include conclusion or answer boilerplate rather than new reasoning content. At each prefix, we append the same answer suffix, ``\texttt{The answer is (}'', and read the next-token logits for the single-token labels A/B/C/D. After a softmax over these four labels, $p_k(\hat{y})$ is the probability assigned, at prefix $k$, to the option the model will eventually output.

The reported ``pre-CoT'' probability is $p_0(\hat{y})$. The lock step is the earliest observed prefix from which the top decoded label is $\hat{y}$ for that prefix and all later prefixes, with a default top-vs.-runner-up probability margin of $0.5$; if no prefix satisfies this condition, the lock point is set to the last observed prefix. Larger lock-step values therefore mean later commitment. The mid-CoT minimum is the minimum of $p_k(\hat{y})$ over the intermediate prefix trajectory.

The observed direction is the opposite of answer-first rationalization. After distillation, the pre-CoT probability of the eventual answer falls from $0.75$ to $0.64$, the mean lock step moves later from $1.35$ to $1.69$, and the mid-CoT minimum falls from $0.69$ to $0.49$. Explicit correction/backtracking markers also become slightly more frequent ($6.8\%{\to}8.2\%$ of steps). The check therefore rules out the simple version of answer-first rationalization: the distilled model is not merely choosing an option before reasoning and then writing a decorative justification for that pre-selected option.

\section{Style and segmentation controls}
\label{app:controls}
The main audit already uses a style-blind prompt, but we also test the two most direct measurement-confound explanations on the first 500 MedQA questions. First, a tone-swap control rewrites the surface form before rejudging: base-model steps are rewritten into fluent prose, and distilled-model steps are rewritten into numbered-outline form. Kimi-K2.6 performs the rewrite, a content-preservation check, and the final style-blind judgment. This makes the control a targeted surface-register test rather than a fully independent rewriter study: the three prompts are separate, and the final judgment prompt sees only the rewritten step, question, and answer key, not the source step or the preservation verdict. We use the preservation-verified subset as the primary analysis. Among preserved rewrites, base-to-prose has $2953$ correct, $1044$ error, and $40$ \texttt{uncertain} judgments ($26.1\%$ error), while distilled-to-outline has $1778$ correct, $1624$ error, and $27$ \texttt{uncertain} judgments ($47.7\%$ error). The paired per-question gap is $+18.6\pp$ with bootstrap $95\%$ CI $[+16.2,+21.0]\pp$ and Wilcoxon $p=4.5{\times}10^{-39}$ over $498$ paired questions. Including all rewrites, regardless of preservation verdict, gives a similar paired gap of $+17.2\pp$ ($95\%$ CI $[+14.8,+19.6]\pp$, $p=4.1{\times}10^{-35}$, $n=500$).

Second, a model-agnostic segmentation control ignores the original numbered or paragraph boundaries and applies the same splitter to both traces. Absolute rates are higher here because each segment can contain multiple factual claims; the comparison of interest is the paired before--after difference under the same splitter. With sentence chunks, base traces have $1025$ correct, $1546$ error, and $48$ \texttt{uncertain} judgments ($60.1\%$ error), while distilled traces have $516$ correct, $1779$ error, and $56$ \texttt{uncertain} judgments ($77.5\%$ error). The paired gap is $+13.4\pp$ ($95\%$ CI $[+10.7,+15.9]\pp$, Wilcoxon $p=4.0{\times}10^{-22}$, $n=500$). With fixed word windows, the corresponding rates are $59.1\%$ vs.\ $77.3\%$, and the paired gap is $+15.2\pp$ ($95\%$ CI $[+12.4,+17.9]\pp$, $p=2.3{\times}10^{-24}$, $n=500$).

\paragraph{Answer-blind judge control.} The main judge sees the answer key so that isolated steps can be interpreted in the context of the correct option. This could bias the role-wise analysis against exploratory hypotheses that point away from the gold answer. We therefore re-audit the same first 500 MedQA questions with Kimi-K2.6 under a prompt that shows the question, options, and one step, but not the correct answer. The prompt explicitly instructs the judge not to penalize a step merely for considering an alternative diagnosis, mechanism, or option, and to mark \texttt{uncertain} when the step cannot be judged without the final answer. The overall gap remains large: $\van$ has $2821$ correct, $1141$ error, and $382$ \texttt{uncertain} judgments ($28.8\%$ committed error), while $\sft$ has $1837$ correct, $1578$ error, and $84$ \texttt{uncertain} judgments ($46.2\%$). The paired per-question gap is $+14.2\pp$ with bootstrap $95\%$ CI $[+11.8,+16.6]\pp$ and Wilcoxon $p=4.2{\times}10^{-26}$ over $491$ paired questions. The hypothesis-generation subset is directionally positive but smaller and underpowered under this stricter prompt ($33.2\%\!\to\!43.3\%$ pooled; paired $+8.0\pp$, $p=0.071$), so we do not rely on that role peak for the main claim. The answer-blind control supports the overall accuracy--reasoning split while narrowing the interpretation of the role-wise peak.

\section{Full per-judge results and judge agreement}
\label{app:judges}
Table~\ref{tab:appjudges} gives the per-judge counts behind the judge-panel summary in the main text. All retained judge configurations show $\sft{>}\van{}$ and (where the teacher was audited) $\teach{}<\van{}$. Inter-judge Cohen's $\kappa$ on shared steps: GLM-4-32B vs.\ Hunyuan-A13B (style-blind) $\kappa{=}0.434$ overall ($n{=}1411$ steps), ``moderate'' agreement on absolute labels. The direction ($\sft{>}\van$) is unanimous across the retained configurations, and restricted to steps both judges call the same way, the distillation inflation is $+9.8\pp$ (cleaner where the judges agree). Per-role $\kappa$ (same judge pair, decomposed by reasoning role): hypothesis-generation $\kappa{=}0.517$ ($n{=}249$), option-elimination $0.524$ ($n{=}126$), factual claim $0.335$ ($n{=}428$), correction $0.297$ ($n{=}107$), structural ``other'' $0.429$ ($n{=}493$); final-synthesis steps are too few in the shared 500-question subset to report a per-role $\kappa$ ($n{<}10$). The judges agree most on hypothesis-generation and option-elimination steps and least on isolated factual claims and corrections; importantly, hypothesis-generation $\kappa$ is \emph{higher} than factual-claim $\kappa$, so the construct-validity concern that exploratory hypotheses are systematically harder to judge as factual is empirically not the case in our data.

\emph{On the high abstention rates of the smaller judges.} The Kimi-K2.6 primary judge declines on $4$--$9\%$ of steps; the smaller GLM-4-32B and Hunyuan-A13B judges decline on $52$--$76\%$. This pattern tracks judge capability: under the strict style-blind prompt, the smaller judges prefer to abstain rather than commit to an uncertain call, while Kimi-K2.6 commits much more often. The abstentions are symmetric across \van{}, \teach{}, \sft{}, and \weakt{} within any one judge (within $\sim$10pp), so they do not differentially favour any condition. On the subset of steps where these smaller judges \emph{do} commit, the distilled--base direction is preserved with pooled effect sizes of $+7.7$ to $+14.3\pp$. We therefore read the GLM and Hunyuan results as a \emph{directional} confirmation on decisively-called steps, not as an absolute-magnitude estimator. The primary magnitude estimate relies on Kimi-K2.6: a paired per-question gap of $+16.1\pp$ ($95\%$ CI $[+14.5,+17.7]\pp$, $1{,}086$ paired questions; Section~\ref{sec:cost}). The non-style-blind GLM-4-32B prompt gives a larger gap than the style-blind one ($+14.3$ vs.\ $+7.7\pp$): confident phrasing is part of what a naive judge counts, but the majority of the inflation survives the style-blind protocol.

\emph{Extreme abstention imputations.} We also recompute the retained judge results under two worst-case conventions: every \texttt{uncertain} step is counted as correct, or every \texttt{uncertain} step is counted as erroneous. For the primary Kimi-K2.6 MedQA audit, the distilled-minus-base gap remains $+18.1\pp$ under the all-correct imputation and $+18.0\pp$ under the all-error imputation. MedMCQA is similarly stable ($+17.3$ and $+16.0\pp$). For the high-abstention smaller judges, the all-correct lower bound is smaller but remains nonnegative: GLM-4-32B style-blind gives $+1.6\pp$ and Hunyuan-A13B style-blind gives $+0.25\pp$; the all-error imputation gives $+9.9$ and $+9.0\pp$. This reinforces the interpretation above: smaller judges are useful for sign checks on committed labels, while Kimi-K2.6 supplies the primary magnitude estimate. The same sensitivity check is reported for GSM8K in App.~\ref{app:math}.

\textbf{Second benchmark.} On MedMCQA (500 questions, Kimi-K2.6 style-blind), $\van$ scores $32.2\%$ ($2486$ correct / $1182$ error) and $\sft$ scores $49.5\%$ ($1410$ / $1382$), a $+17.3\pp$ inflation, while greedy accuracy is $59.6\%$ vs.\ $59.4\%$ (the MedQA-trained student is no more accurate on MedMCQA).

\textbf{The hedge-band-freeze experiment} (setup in App.~\ref{app:hedge}). On a 500-question subset (smaller-scale separate training runs), the step-error rate is $20.8/18.4/20.4\%$ under GLM-4-32B (s.b.)\ and $44.3/43.7/44.8\%$ under Kimi-K2.6 for the (freeze-nothing / freeze-hedge-band / freeze-control-band) conditions: a $\sim$2\pp{} improvement under the lenient judge but within noise under the strict one, so we treat the channel as a read-out, not a causal lever.

\textbf{Clinical expert audit.} Two clinical experts independently labelled the same 450 steps, stratified by model, reasoning role, and judge label: 150 per model and 30 per model--role cell. Model identities and judge labels were hidden. The sample excludes judge-\texttt{uncertain} and structural ``other'' steps. Within each model, we post-stratify role $\times$ judge-label cells to their eligible-population frequencies and divide weighted human errors by weighted human non-\texttt{uncertain} labels. The distilled-minus-base gaps are $+13.0\pp$ and $+12.9\pp$ for experts 1 and 2; both rank $\teach{<}\van{<}\sft$. Three-label agreement is $81.6\%$ (Cohen's $\kappa=0.556$, $n=450$); excluding steps either expert marked \texttt{uncertain}, $\kappa=0.589$ ($n=440$). Among steps each expert labelled erroneous, medically false or clinically critical errors account for $41/47$ base and $48/53$ distilled steps for expert~1, and $44/46$ and $54/56$ for expert~2. These results support the direction and substantive nature of the shift within the sampled roles, rather than an all-step medical error prevalence.

\begin{table}[t]
\centering\footnotesize\setlength{\tabcolsep}{2pt}
\begin{tabular}{@{}llrrrr@{}}
\toprule
Judge & Cond. & \#cor & \#err & \#unc & err\,\% \\
\midrule
\multirow{4}{*}{Kimi-K2.6 (s.b.)} & \van & 7081 & 3117 & 997 & 30.6 \\
 & \teach{} & 5700 & 1194 & 301 & 17.3 \\
 & \sft & 4058 & 4113 & 791 & 50.3 \\
 & \weakt & 4749 & 3312 & 809 & 41.1 \\
\midrule
\multirow{4}{*}{GLM-4-32B (s.b.)} & \van & 1563 & 167 & 2685 & 9.7 \\
 & \teach & 920 & 68 & 2229 & 6.9 \\
 & \sft & 887 & 186 & 2409 & 17.3 \\
 & \weakt & 919 & 170 & 2384 & 15.6 \\
\midrule
\multirow{4}{*}{Hunyuan-A13B (s.b.)} & \van & 1183 & 264 & 2968 & 18.2 \\
 & \teach & 682 & 99 & 2436 & 12.7 \\
 & \sft & 619 & 217 & 2646 & 26.0 \\
 & \weakt & 623 & 223 & 2627 & 26.4 \\
\midrule
\multirow{4}{*}{GLM-4-32B (orig.)} & \van & 1854 & 244 & 2317 & 11.6 \\
 & \teach & 1234 & 84 & 1899 & 6.4 \\
 & \sft & 1124 & 394 & 1964 & 26.0 \\
 & \weakt & 1169 & 288 & 2016 & 19.8 \\
\bottomrule
\end{tabular}
\caption{Step counts (correct\,/\,error\,/\,uncertain) and error rate (\%) per judge. Kimi-K2.6 over the full 1{,}273-question set; GLM-4-32B and Hunyuan-A13B over a random 500-question subset.}
\label{tab:appjudges}
\end{table}

\section{MedBullets5 clinical-vignette check}
\label{app:medbullets}
MedBullets5 is a clinical-vignette benchmark with five answer options and expert-written explanations \citep{chen-etal-2025-benchmarking}. The public five-option release has $308$ rows but only $298$ unique question texts, because ten rows duplicate earlier questions. We audit the $298$ unique questions so that a duplicated case cannot influence the rate twice.

For this transfer check, $\van$ and $\sft$ are the same Qwen3-8B base and MedQA-distilled student used elsewhere; $\teach$ is a fresh DeepSeek-V3.2 non-thinking teacher trace generated for the same questions. Because the traces vary more in surface format than the primary MedQA traces, we use the same uniform sentence-chunk splitter for all three conditions. This makes the absolute rates a segment-level estimate rather than directly comparable to the numbered-step MedQA rate. The comparison of interest is still within-condition and paired: the same questions, splitter, prompt, and judge are used for all three traces.

\begin{table}[h]
\centering\scriptsize
\setlength{\tabcolsep}{2pt}
\begin{tabular}{@{}lrrrrr@{}}
\toprule
Condition & Acc. (\%) & \#cor & \#err & \#unc & Err. (\%) \\
\midrule
Teacher & 76.5 & 519 & 412 & 0 & 44.3 \\
Base & 53.4 & 587 & 1112 & 7 & 65.5 \\
Distilled & 57.4 & 314 & 1137 & 0 & 78.4 \\
\bottomrule
\end{tabular}
\caption{MedBullets5 clinical-vignette transfer check on $298$ unique questions. Accuracy is single-chain answer accuracy for the audited trace. Segment-error rate is Kimi-K2.6 style-blind error/(error+correct) under a uniform sentence-chunk splitter.}
\label{tab:medbullets}
\end{table}

The ordering is the same as in the primary medical audit, despite the different benchmark and splitter. The teacher is strongest at the answer level and cleanest at the process level. The distilled student improves the audited-chain answer accuracy by $+4.0\pp$ over the base student ($53.4\%\!\to\!57.4\%$), but its segment-error rate rises from $65.5\%$ to $78.4\%$. Paired by question, the distilled-minus-base increase is $+10.0\pp$ with bootstrap $95\%$ CI $[+6.7,+13.2]\pp$ and Wilcoxon $p=1.3{\times}10^{-11}$ ($195$ questions worse, $69$ better, $34$ tied). Relative to the teacher, the distilled trace is higher by $+33.3\pp$ on average ($95\%$ CI $[+29.4,+37.1]\pp$).

\section{Math-domain diagnostics}
\label{app:math}
The GSM8K check uses a separate Qwen3-8B student fine-tuned on GSM8K teacher CoTs and audited on the full 1{,}319-question GSM8K test set. We keep the style-blind audit structure but change the target from medical factuality to computational and logical correctness. The judge sees the word problem, the gold numeric answer, and one step, and labels only arithmetic errors, invalid algebraic manipulations, misapplied formulae, or non-following logical inferences.

For answer accuracy we use a final-number parser that accepts the generated formats used by both models: \texttt{\textbackslash boxed\{\}}, ``final answer'' / ``the answer is'' patterns with intervening markdown, and, if neither appears, the last numeric expression near the end of the trace. This corrects a formatting artifact in the raw generation logs, where many base-model answers such as ``The answer is **18**'' were marked wrong by a stricter extractor.

\begin{table}[t]
\centering\scriptsize
\setlength{\tabcolsep}{2pt}
\begin{tabular}{@{}lrrrrrr@{}}
\toprule
Condition & Acc. (\%) & \#cor & \#err & \#unc & Err. (\%) & Unc. (\%) \\
\midrule
Base & 92.4 & 10822 & 781 & 80 & 6.7 & 0.7 \\
Teacher & 95.1 & 7767 & 754 & 86 & 8.8 & 1.0 \\
Math distilled & 92.6 & 6415 & 818 & 1125 & 11.3 & 13.5 \\
\bottomrule
\end{tabular}
\caption{GSM8K math-domain diagnostic. Accuracy is robust final-number accuracy on the 1{,}319-question test set for base and distilled and $95.1\%$ teacher correctness on the $1{,}318$ teacher-generated test traces available locally. Step-error rate is Kimi-K2.6 style-blind error/(error+correct); \texttt{uncertain} is excluded from the error denominator.}
\label{tab:math}
\end{table}

The answer-level difference between base and distilled students is negligible (paired McNemar $p=0.81$; bootstrap $95\%$ CI for the accuracy change $[-1.1,+1.4]\pp$). The pooled committed step-error rate rises by $+4.6\pp$ ($6.7\%\!\to\!11.3\%$), and over the $1{,}145$ questions with at least one committed step in both arms, the paired per-question increase is $+3.4\pp$ with bootstrap $95\%$ CI $[+1.9,+5.1]\pp$ and Wilcoxon $p=2.2{\times}10^{-7}$. But the teacher trace is already noisier than the base trace ($8.8\%$ vs.\ $6.7\%$), and a 500-question MATH subset \citep{hendrycks2021math} shows the same teacher/base mismatch. These math checks therefore violate the clean-teacher premise of the primary result and should be read as boundary diagnostics, not as positive cross-domain replications. On GSM8K, distilled traces are shorter on average (6.3 vs.\ 8.9 audited steps), so the higher step-error rate is not produced by longer generated solutions. The \texttt{uncertain} asymmetry is real: the distilled math traces have a $13.5\%$ abstention rate vs.\ $0.7\%$ for the base. Counting every \texttt{uncertain} step as correct still leaves a $+3.1\pp$ gap; counting every one as erroneous gives $+15.9\pp$.

\section{Science-domain near-ceiling boundary}
\label{app:arc}
To probe the opposite limit from the weak-base boundary, we ran the full distillation-and-audit procedure on ARC-Challenge, a multiple-choice grade-school science benchmark. A separate Qwen3-8B student was fine-tuned on DeepSeek teacher CoTs for $3{,}303$ teacher-correct training questions (same non-thinking teacher and first-round teacher-correct filter as the primary setting, without the answer-hinted second round), then base and distilled students each generated one chain for the full $1{,}171$-question Challenge test set, audited with the same Kimi-K2.6 style-blind prompt adapted to general scientific rather than medical content.

ARC lacks two ingredients of the medical risk regime: answer headroom is small, and the rationale is short and tightly tied to the selected option. The Qwen3-8B base already answers $92.6\%$ correctly, and the distilled student reaches $94.6\%$, a $+2.0\pp$ change. The step-error summaries are correspondingly small and weighting-sensitive. Pooled over committed steps, error moves from $9.2\%$ for the base ($658/7142$ committed steps) to $8.6\%$ for the distilled student ($472/5483$), a $-0.6\pp$ gap with bootstrap $95\%$ CI $[-1.75,+0.49]\pp$. Averaged by question, the sign flips to $+1.1\pp$ (Wilcoxon $p=0.036$; $240$ questions worse, $205$ better, $726$ tied), because distilled chains are shorter (4.68 vs.\ 6.10 committed steps per question) and per-question weighting gives short degraded chains more influence. Both summaries are at the $\sim$1pp scale, far from the $+17$--$20\pp$ pooled increases on MedQA and MedMCQA. Abstentions are rare in both arms ($5$ vs.\ $2$ \texttt{uncertain} steps). We therefore read ARC neither as a reversal nor as a replication, but as a boundary case with no large, weighting-stable trace degradation.

\section{Student-side regime checks}
\label{app:families}
We ran the same 500-question generation-and-audit procedure for four student-side checks distilled from the same teacher-generated CoT style. Qwen3-14B and Qwen3-32B test whether the split survives larger same-family students. Llama-3.1-8B tests a second capable 8B family. Mistral-7B locates the weak-student boundary. Table~\ref{tab:families} reports both the answer-level change and the step-level audit change under the same Kimi-K2.6 style-blind judge. The larger Qwen3 effects are smaller than the primary Qwen3-8B effect but statistically positive in paired analysis; Llama repeats the larger 8B degradation. Mistral-7B reverses the main pattern from a weaker starting point and with less controlled trace segmentation: only $2.6/13.4\%$ of base/distilled Mistral chains have explicit numbered-step boundaries, compared with $22.2/11.2\%$ for Qwen3-8B and $68.2/11.2\%$ for Llama. We interpret this formatting instability as part of the weak-base regime: the same weakness that gives supervised fine-tuning room to improve answers and traces also makes the raw traces less regularly structured. This is why we use Mistral as the weak-base boundary condition in the main text rather than as an equally clean estimate of the capable-student effect.

\begin{table*}[t]
\centering\footnotesize
{\setlength{\tabcolsep}{0pt}
\renewcommand{\arraystretch}{1.05}
\begin{tabular*}{0.82\textwidth}{@{\extracolsep{\fill}}lrrrrrr@{}}
\toprule
& \multicolumn{2}{c}{Answer acc. (\%)} & \multicolumn{2}{c}{Step error (\%)} & \multicolumn{2}{c}{Uncertain steps (\%)} \\
\cmidrule(lr){2-3}\cmidrule(lr){4-5}\cmidrule(l){6-7}
Student & Base & Distilled & Base & Distilled & Base & Distilled \\
\midrule
Qwen3-8B & 71.6 & 74.4 & 29.7 & 49.7 & 2.2 & 0.5 \\
Qwen3-14B & 74.6 & 80.8 & 31.5 & 37.9 & 0.8 & 0.5 \\
Qwen3-32B & 81.8 & 84.2 & 22.8 & 30.6 & 2.5 & 0.6 \\
Llama-3.1-8B & 66.8 & 73.6 & 31.2 & 45.5 & 0.1 & 0.4 \\
Mistral-7B & 48.6 & 67.6 & 56.5 & 49.1 & 0.1 & 14.9 \\
\bottomrule
\end{tabular*}}
\caption{Student-side regime checks on random 500-question MedQA subsets. Accuracy is answer accuracy from one generated CoT per question. Step-error rate is judged by Kimi-K2.6 with the same style-blind prompt used in the primary audit. ``Uncertain steps'' is the fraction of step judgments where the judge abstains; these steps are excluded from the step-error denominator.}
\label{tab:families}
\end{table*}

\section{Trace-function taxonomy}
\label{app:taxonomy}
Each step is assigned a single role by a priority-ordered keyword classifier: \emph{correction / backtracking} (``however'', ``wait'', ``reconsider'', ``on second thought'', \ldots), then \emph{final synthesis} (``the answer is'', ``in conclusion'', ``therefore'', \ldots), then \emph{option elimination} (``rule out'', ``less likely'', ``option A/B/C/D'', ``(A)'', \ldots), then \emph{hypothesis generation / differential diagnosis} (``differential'', ``could be'', ``consider'', ``most likely diagnosis'', \ldots), then \emph{other} (structural headers, $<$25 characters, ``let me work through this'', \ldots), and otherwise \emph{factual claim} (the default---a step stating a medical fact, mechanism or value). ``Exploratory'' aggregates hypothesis-generation, option-elimination and correction. The distilled model writes far fewer markdown-header ``other'' steps and correspondingly more prose ``factual claim'' steps---a segmentation artifact that does not by itself explain the within-role error-rate increases in Section~\ref{sec:taxonomy} or the unchanged count of erroneous exploratory steps per chain. Because the classifier is keyword-based, style shifts can still move borderline steps between roles; we therefore use the taxonomy as a localization analysis rather than as the main estimate. The robust observation is that error rates rise within \emph{every} reasoning-bearing role, while App.~\ref{app:judges} reports that judges agree at least as well on hypothesis-generation and option-elimination steps as on isolated factual claims.

\section{Hedge-direction localisation}
\label{app:hedge}
Section~\ref{sec:fix} uses projection-out ablation to locate the base model's hedge-token read-out and patch recovery to compare base--distilled differences across model families. These experiments test whether the visible caution channel survives distillation despite failing to track the new trace-error baseline.

\paragraph{Hedge direction.} Following \citet{arditi2024refusal}, for each layer $\ell$ we form $\hat d_\ell$ as the unit-normalised difference of means between two sets of last-token residual-stream activations from base-model CoTs. The projection-out experiment uses no base--distilled disagreement threshold. A high-hedge position is a prefix where the base model assigns at least $0.05$ next-token probability mass to the hedge-marker set below; a low-hedge control position is a prefix where that mass is below $0.01$. We consider positions after token 20, truncate prefixes at 220 tokens, take the top $80$ high-hedge positions by hedge mass, and sample $160$ low-hedge controls uniformly from the collected pool.

The hedge-marker set is the single-token encodings, under the relevant model tokenizer, of the following space-prefixed strings: \texttt{might}, \texttt{may}, \texttt{could}, \texttt{possibly}, \texttt{perhaps}, \texttt{probably}, \texttt{likely}, \texttt{unlikely}, \texttt{suggest}, \texttt{suggests}, \texttt{indicate}, \texttt{indicates}, \texttt{appears}, \texttt{appear}, \texttt{consistent}, \texttt{typically}, and \texttt{generally}. The specificity controls use the same single-token rule for certainty markers (\texttt{definitely}, \texttt{certainly}, \texttt{clearly}, \texttt{obviously}, \texttt{surely}, \texttt{indeed}, \texttt{confirmed}, \texttt{established}, \texttt{precisely}, \texttt{exactly}, \texttt{absolutely}) and first-person markers (\texttt{I}, \texttt{we}, \texttt{my}, \texttt{our}, \texttt{us}, \texttt{let's}, \texttt{Let's}). At inference we project the direction out of the residual stream, $x\leftarrow x-(x\cdot\hat d_\ell)\hat d_\ell$, at a chosen set of layers. Hedge-token probability is the next-token probability mass assigned to the hedge-marker set.

\begin{figure*}[t]
\centering
\includegraphics[width=\textwidth]{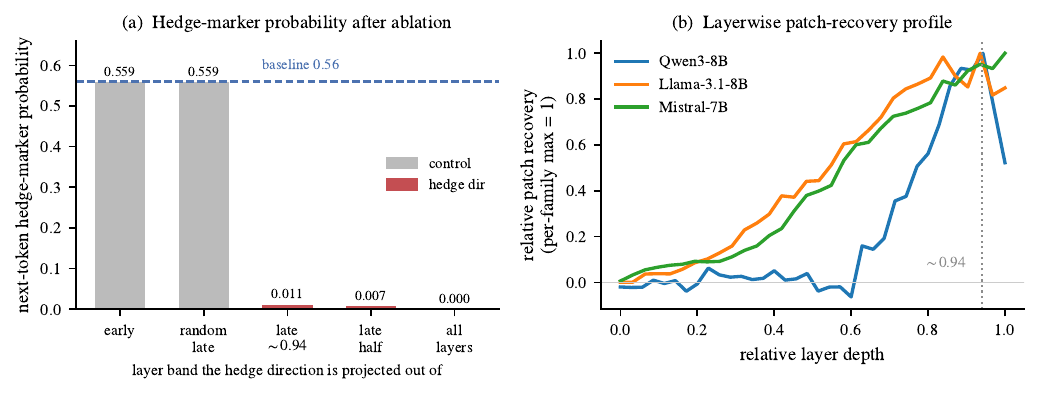}
\caption{\textbf{(a)} Projection-out ablation of the hedge direction. Bars show absolute next-token probability mass on the fixed hedge-marker list. Early-layer and random late-band controls stay near the unablated baseline (dashed line), while projecting out the fitted direction at relative depth $\approx 0.94$ drives hedge-token probability from about $0.56$ to $0.011$ (about a $98\%$ drop). Broader late-half and all-layer ablations reach the same ceiling. \textbf{(b)} Layerwise patch-recovery profile. Recovery peaks in the last $\sim$10\% of layers in all three model families. The plotted value is $R_\ell/\max_{\ell'}R_{\ell'}$, so $1$ marks the peak layer for that family rather than full restoration; pairwise curve correlations are Pearson $r=0.81$--$0.98$.}
\label{fig:channel}
\end{figure*}

The about-$98\%$ value in Figure~\ref{fig:channel}a is the full-set directional ablation: fit $\hat d_\ell$ from all $80$ high-hedge and all $160$ low-hedge positions, then evaluate on the same high-hedge positions. The held-out check splits the same counts into $50/30$ high-hedge train/test positions and $100/60$ low-hedge train/test controls. We fit the direction on the train split and measure the hedge-probability drop only on the $30$ held-out high-hedge positions, because the target quantity is loss of hedge mass where hedging was available. This gives a $99.7\%$ drop. The two percentages are not meant to rank full-set vs.\ held-out performance; they come from different high-hedge evaluation subsets and both indicate the same near-total ablation pattern. The unused low-hedge test controls are kept only to make the split symmetric.

Controls: (i) the same direction projected out at early layers (rel.\ $0.08$--$0.14$); (ii) a random unit direction projected out at the late band (5 seeds); (iii) the same ablation measured on certainty-marker and first-person-marker probabilities. An alternative direction estimator, a supervised logistic-regression hyperplane separating high-hedge from low-hedge contexts, is fit on the same train split and evaluated on the same held-out high-hedge positions. It yields a much smaller $28.6\%$ ablation drop. We use this only as a specificity check: the near-total ablation is not a generic consequence of removing any linear high-hedge/low-hedge separator. It does not prove that the difference-of-means direction is unique or that it is a complete circuit.

\paragraph{Patch recovery.} Figure~\ref{fig:channel}b uses a different intervention from the projection-out ablation in Figure~\ref{fig:channel}a. For each high-hedge prefix, we run the distilled model and replace its last-token residual stream at one layer with the base model's residual stream at the same prefix. The patched value is the hedge-token probability after this base-to-distilled replacement. Recovery is
\[
R = \frac{p_{\mathrm{patch}} - p_{\mathrm{dist}}}
         {p_{\mathrm{base}} - p_{\mathrm{dist}}},
\]
where $p_{\mathrm{base}}$ and $p_{\mathrm{dist}}$ are the unpatched base and distilled hedge-token probabilities on the same prefixes.
A recovery value of $0$ means the patch did not move the distilled model toward the base model's hedge probability; $1$ means it restored the full base--distilled gap. We compute the analogous per-layer quantity for Llama-3.1-8B and Mistral-7B under their own distillation runs and align layers by relative depth.
Figure~\ref{fig:channel}b plots $R_\ell / \max_{\ell'} R_{\ell'}$ separately for each family, so the plotted value $1$ marks that family's strongest recovery layer rather than full restoration.

\paragraph{The hedge-band-freeze runs.} The freeze-hedge-band run freezes Qwen3-8B layers $30$--$34$ during LoRA fine-tuning. This is a coarse training-time protection band around the projection-ablation band, layers $31$--$33$, with one neighbouring layer on either side; it should not be read as a sharper localisation claim. The control-band run freezes the equally sized layer range $15$--$19$, and the freeze-nothing run uses the same data and recipe with no frozen transformer blocks. These are separate, smaller-scale training runs from the main \sft{} model. The small hedge-band improvement under GLM-4-32B is within noise under Kimi-K2.6, so we do not treat the band as a causal lever; per-judge numbers are in Appendix~\ref{app:judges}.

\section{Internal step-correctness probe}
\label{app:probe}
For each audited step-end token, we extract residual-stream activations at every layer, apply a fixed Gaussian random projection ($4096{\to}512$; Johnson--Lindenstrauss, no fitting), $z$-score using training-fold statistics, and fit a logistic-regression probe for the judge's \texttt{error} label with 5-fold cross-validation grouped by question.

On the original GLM-4-32B-labelled 500-question audit, peak AUROC is $0.71$ in $\van$ (relative depth $\approx 0.57$) and $0.73$ in $\sft$ ($\approx 0.34$); the weak-teacher and band-frozen variants score $0.72$ and $0.73$, respectively. Repeating extraction and probing over the full 1{,}273-question CoTs with the primary Kimi-K2.6 labels yields the $\approx 0.82$ AUROC cited in the main text (Table~\ref{tab:probekimi}). Absolute AUROC depends on the judge label set and class balance. Pre/post differences are small and change sign across label sets, so we do not interpret them as a meaningful change in internal decodability.

\begin{table}[htbp]
\centering\small
\setlength{\tabcolsep}{4pt}
\begin{tabular}{@{}lccc@{}}
\toprule
Condition & best AUROC & best rel. & late-band AUROC \\
\midrule
\van & 0.832 & 0.66 & 0.815 \\
\sft & 0.819 & 0.63 & 0.808 \\
\weakt & 0.802 & 0.63 & 0.774 \\
\bottomrule
\end{tabular}
\caption{Step-error probe AUROC against Kimi-K2.6 labels on all 1{,}273 questions (5-fold CV grouped by question). ``rel.'' gives the best layer's relative depth; late-band AUROC averages over the hedge-direction band at relative depths $\sim$0.86--0.94 (Figure~\ref{fig:channel}).}
\label{tab:probekimi}
\end{table}

\textbf{Controls.} Within-question label shuffling yields AUROC $\approx 0.60$ rather than $0.50$, indicating nuisance structure. A probe predicting step \emph{position} (early vs.\ late) from the same residuals reaches AUROC $\approx 0.96$. Since later steps are more error-prone (in base traces, step error rises from $8.9\%$ in the first fifth of a chain to $46.0\%$ in the last), position and surface correlates partly explain the $\approx 0.82$ error-probe AUROC. We therefore make only the comparative claim that this linear decodability is not erased by distillation, without attributing it to a dedicated error representation or claiming that the model knows which steps are wrong.

\section{Training details}
\label{app:training}
Table~\ref{tab:training} lists the training settings. The Qwen3-8B primary student, the weaker-teacher and unhinted controls, and the Qwen3-14B and Qwen3-32B students all use this recipe; the larger students use per-device batch size $1$ with $16$ accumulation steps, which keeps the effective batch size at $16$. The Llama-3.1-8B-Instruct and Mistral-7B-Instruct-v0.3 students (App.~\ref{app:families}) and the three Qwen3-8B hedge-band-freeze runs (App.~\ref{app:hedge}) train on the same $20{,}356$ traces with LoRA rank $32$ ($\alpha{=}64$) for two epochs, with all other settings unchanged. The ARC-Challenge student (App.~\ref{app:arc}) uses learning rate $5{\times}10^{-5}$, LoRA dropout $0.1$, and effective batch size $8$. Each student is trained once (seed $42$).

\begin{table}[h]
\centering\footnotesize
\begin{tabular}{@{}lL{0.52\columnwidth}@{}}
\toprule
Setting & Value \\
\midrule
Training questions & $10{,}178$ (MedQA-USMLE train) \\
Traces per question & 2 teacher-correct \\
Teacher sampling & temp.\ $0.7$, top-$p$ $0.95$, ${\le}2{,}048$ tokens \\
Hint rule, DeepSeek & after 5 consecutive wrong answers \\
Hint rule, Qwen3-14B & ${<}2$ of 8 samples correct (temp.\ $0.3$) \\
Training traces & $20{,}356$ ($19{,}486$ unhinted retrain) \\
Held-out split & $2\%$ of traces, not used for selection \\
LoRA & rank 16, $\alpha{=}32$, dropout $0.05$ \\
LoRA targets & q, k, v, o, gate, up, down \\
Optimizer & AdamW, lr $2{\times}10^{-5}$, cosine, $5\%$ warmup \\
Weight decay / grad.\ clip & $0$ / $1.0$ \\
Effective batch / epochs & 16 / 3 \\
Max sequence length & $3{,}072$ tokens \\
Loss & trace tokens only (prompt masked) \\
Chat template & Qwen3 thinking mode disabled \\
Precision & bf16 \\
Hardware & NVIDIA A100, 1--4 GPUs per run \\
Wall-clock per run & 7--8\,h (Qwen3-8B), 11\,h (14B), 23\,h (32B) \\
\bottomrule
\end{tabular}
\caption{Training settings for the MedQA students.}
\label{tab:training}
\end{table}

\end{document}